\documentclass[runningheads]{llncs}

\usepackage{eccv}

\usepackage{eccvabbrv}

\usepackage{graphicx}
\usepackage{booktabs}

\usepackage[accsupp]{axessibility}  

\usepackage{hyperref}
\usepackage{orcidlink}

\usepackage{booktabs} 
\usepackage{multirow} 
\usepackage{amsmath}  
\usepackage{graphicx} 
\usepackage[table]{xcolor} 
\usepackage{graphicx}
\usepackage{xcolor}       
\usepackage{booktabs}
\usepackage{amsmath}
\usepackage{graphicx}   
\usepackage{array}      
\usepackage{tabularx}
\usepackage{tcolorbox}
\usepackage{algorithm}
\usepackage{algpseudocode}
\usepackage{color, colortbl}
\usepackage{booktabs}   
\usepackage{multirow}   
\usepackage{amsmath}    
\usepackage{amssymb}    
\usepackage{pifont}     
\usepackage{tabularx}
\newcommand{\cmark}{\ding{51}}
\newcommand{\xmark}{\ding{55}}

\definecolor{myteal}{HTML}{428989}   
\definecolor{mypink}{HTML}{DB2777}   
\definecolor{myblue}{HTML}{0000FF}   

\makeatletter
\def\@fnsymbol#1{\ensuremath{\ifcase#1\or *\or \dagger\or \ddagger\or
   \mathsection\or \mathparagraph\or \|\or **\or \dagger\dagger
   \or \ddagger\ddagger \else\@ctrerr\fi}}
\makeatother

\begin{document}

\title{PixelU: A U-Shaped Transformer for Efficient End-to-End Pixel Diffusion} 

\titlerunning{PixelU}

\author{
Zipeng Guo\textsuperscript{1}\thanks{Equal Contribution.} \and
Lichen Ma\textsuperscript{1}\textsuperscript{2}$^*$\thanks{Project Lead.} \and
Yu He\textsuperscript{1}$^*$ \and
Xiaolong Fu\textsuperscript{1} \and
Jingling Fu\textsuperscript{1} \and
Junshi Huang\textsuperscript{1}\thanks{Corresponding Author.} \and
Yan Li\textsuperscript{1}
}

\authorrunning{Z. Guo et al.}

\institute{
\textsuperscript{1}JD.com, \textsuperscript{2}Xi'an Jiaotong University\\
\email{\{guozp8888, malichen2020, junshi.huang\}@gmail.com}
}

\maketitle

\begin{abstract}

End-to-end pixel-space diffusion models bypass the lossy compression of Latent Diffusion Models (LDMs) but struggle to jointly model low-frequency semantics and high-frequency signals in high-dimensional space. Existing works heavily rely on complex pixel decoders to alleviate this issue. In this paper, we challenge this trend by revealing that these decoders primarily compensate for the optimization difficulties inherent to velocity prediction ($v$-prediction). Under the clean data paradigm ($x$-prediction), they are redundant. Motivated by this insight, we advocate for simplicity over complexity and introduce PixelU, a minimalist, single-stage U-shaped Diffusion Transformer tailored for pixel space. PixelU abandons auxiliary decoders in favor of zero-cost skip connections, which provide an ``information highway'' that directly routes uncorrupted high-frequency spatial details from shallow to deep layers. To further enable the backbone to focus exclusively on modeling low-frequency semantics, we introduce a constant-channel spatial down-sampling mechanism as a natural low-pass filter, which compresses deep features into a compact, low-frequency semantic manifold. Extensive experiments demonstrate that this decoupling of frequencies could outperform the strong baseline (JiT-G) with only about $1/3$ of its computation cost. On ImageNet $256\times256$ and $512\times512$, PixelU achieves FID of $1.63$ and $1.92$ respectively, surpassing recent pixel-space methods and establishing a simple yet powerful new paradigm for end-to-end diffusion models. Code will be available at \href{https://github.com/gzp6688/PixelU}{https://github.com/gzp6688/PixelU}.

\end{abstract}

\section{Introduction}
\label{sec:intro}

The remarkable success of contemporary image generation is largely driven by Latent Diffusion Models (LDMs) \cite{rombach2022high,ma2024sit,peebles2023scalable,esser2024scaling}. By compressing raw pixels into a latent space via pre-trained VAEs, LDMs effectively filter out high-frequency spatial details to simplify the generation process. However, this reliance on external VAEs inherently limits fine-grained image fidelity and restricts the architecture to a two-stage pipeline. Consequently, there is a surging interest in end-to-end pixel-space diffusion models, which bypass VAEs to synthesize images directly in the high-dimensional raw pixel space.

Unlike LDMs, pixel-space diffusion \cite{chen2025pixelflow,dhariwal2021diffusion,hoogeboom2023simple,kingma2023understanding,teng2023relay} presents a severe optimization challenge: a single diffusion transformer (DiT) \cite{peebles2023scalable} struggles to simultaneously model low-frequency global semantics and high-frequency signals (e.g., high-dimensional noise and fine image details like edges and textures). To alleviate this, a series of recent state-of-the-art works \cite{wang2025pixnerd,chen2025dip,ma2025deco,yu2025pixeldit} have attempted to preserve high-frequency spatial details by adding architecturally complex pixel decoders. While effective, these specialized modules incur significant computational costs that degrade the overall training and inference efficiency. In stark contrast, recent JiT \cite{li2025jit} demonstrate that remarkable generation quality can be achieved on a plain DiT simply by shifting the prediction objective to clean data ($x$-prediction) rather than velocity ($v$-prediction).

This leads us to rethink the prevailing architectural trend and raises a critical question: \textit{Are such heavyweight decoders truly necessary for high-frequency modeling in pixel space?} To answer this, we conduct a systematic evaluation of these decoders from the perspective of prediction objectives. Our empirical analysis reveals a crucial insight: \textit{complex pixel decoders primarily compensate for the optimization difficulties inherent to $v$-prediction}. Under the $x$-prediction paradigm, which naturally anchors the generative target on a low-dimensional image manifold, these decoders become largely redundant, offering only marginal performance gains at massive computational costs.

Motivated by this insight, we advocate for simplicity over complexity. Rather than relying on complex pixel decoders to recover fine details, we revert to a classic, essentially zero-cost architectural design: \textbf{Skip Connections}. Since $x$-prediction explicitly targets the clean image, skip connections naturally provide the exact ``information highway'' needed. They directly route uncorrupted high-frequency spatial details from the shallow encoder straight to the decoder, completely replacing the need for complex pixel-wise modeling at the end of the network. Having delegated the preservation of high-frequency details to the skip connections, the backbone network is liberated to focus exclusively on low-frequency global semantics. By visualizing the intermediate predictions $\hat{x}_0$ across the diffusion trajectory, we observe that the generative process intrinsically prioritizes macroscopic low-frequency structures in its early stages. This trajectory demonstrates that before resolving complex high-dimensional details, the core structure of an image is first constructed on a compact, \textit{endogenous semantic manifold} governed by low-frequency signals. To explicitly compel the network to capture this manifold, we introduce \textbf{Spatial Down-sampling}. Functioning as a natural, physical low-pass filter, down-sampling systematically attenuates high-frequency noise \cite{williams2023unified,tian2024u} and compresses entangled features into compact, semantically rich representations, thereby accelerating convergence.

Integrating these insights, we introduce \textbf{PixelU}, a minimalist, single-stage U-shaped Diffusion Transformer tailored for pixel space. Unlike traditional U-Net architectures \cite{ronneberger2015u,ho2020denoising,rombach2022high}, PixelU performs only a single stage of spatial down- and up-sampling and maintains a constant channel dimension, achieving massive computational savings while yielding outstanding generative quality.

Extensive experiments show that PixelU generates high-quality images when trained end-to-end on pixel space without any autoencoders. On ImageNet $256\times256$ and $512\times512$, PixelU achieves FID of $1.63$ and $1.92$ respectively, outperforming recent pixel-space models by a large margin. Notably, it surpasses the strong JiT-G baseline \cite{li2025jit} while requiring merely 1/3 of the computational cost. We would like to highlight that our objective is not to reinvent the U-Net, but to offer a timely reminder to the community. Amidst the recent trend toward flat DiTs \cite{peebles2023scalable,ma2024sit} and complex pixel decoders \cite{wang2025pixnerd,chen2025dip,ma2025deco,yu2025pixeldit}, the foundational design of skip connections and spatial down-sampling has been largely overlooked. We demonstrate that under the $x$-prediction objective, these two classic components synergistically decouple high-frequency details from low-frequency semantics, establishing an elegant and compute-friendly new paradigm for pixel diffusion.

\section{Related Work}

\subsection{Latent Space Diffusion}
Latent diffusion models (LDMs) compress raw images into a compact latent space via pre-trained Variational Autoencoders (VAEs) and subsequently train the diffusion models within this space \cite{rombach2022high}. 
Early LDMs primarily relied on CNN-based U-Net architectures \cite{dhariwal2021diffusion,ho2020denoising,song2020denoising,song2021score}, whereas the pioneering work of DiT \cite{peebles2023scalable} introduced a flat, isotropic Transformer architecture to replace the U-Net.

However, recent studies have begun to re-evaluate the utility of the classic U-Net macro-architecture in the Transformer era. For instance, U-ViT \cite{bao2023uvit} designs a ViT-based U-shaped architecture that assists diffusion generation by incorporating long skip connections between shallow and deep layers. U-DiT \cite{tian2024udit} further identifies potential redundancies in purely isotropic DiTs, introducing token downsampling within the self-attention mechanism of a U-shaped diffusion Transformer. To further enhance the accuracy of generated semantics and accelerate convergence, studies such as REPA \cite{yu2025repa} leverage pre-trained large vision models to align intermediate representations. Addressing the spatial dimension inconsistencies caused by downsampling in U-Net architectures, U-REPA \cite{tian2025urepa} proposes a specialized alignment paradigm, successfully extending REPA to U-Nets. DDT \cite{wang2025ddt} explores single-scale frequency decoupling in a compressed latent space, showing that frequency decoupling remains important even in compressed space. Despite these advancements, the two-stage paradigm is inherently bottlenecked by VAEs. Their lossy compression inevitably introduces decoding artifacts and obliterates high-frequency details, imposing a strict upper bound on generation quality and image fidelity \cite{rombach2022high,chen2025deep}.


\subsection{Pixel Space Diffusion}
To overcome the numerous limitations imposed by VAEs, pixel-space diffusion models advocate for direct, end-to-end modeling in the raw pixel space. Early diffusion models performed denoising directly on pixels, but constrained by the enormous dimensionality, they faced immense challenges regarding computational burden and optimization difficulty. To reduce the computational overhead of high-resolution image generation, previous research explored multi-scale Cascaded Diffusion \cite{ho2022cascaded} or Relay Diffusion \cite{teng2023relay}. These approaches divide the generation process into multiple resolution stages or utilize multi-scale architectures (e.g., PixelFlow \cite{chen2025pixelflow}). However, they often incur higher training costs and introduce complex inference scheduling.

Recent works are dedicated to advancing pixel-space generation by improving architectures and prediction targets. In terms of architectural innovation, FractalGen \cite{li2025fractal} constructs a fractal generative model with long-range structures, TarFlow \cite{zhai2024tarflow} and FARMER \cite{zheng2025farmer} integrate Transformers with normalizing flows to directly model pixels, while PixNerd \cite{wang2025pixnerd} utilizes lightweight neural field layers for pixel-space diffusion. Regarding improvements in prediction targets, JiT \cite{li2025jit} demonstrates that shifting the training objective from velocity prediction to directly predicting the clean image ($x$-prediction) can effectively simplify the learning process in high-dimensional spaces and anchor the generation on a low-dimensional image manifold.

Furthermore, several recent works \cite{ma2025deco,chen2025dip,yu2025pixeldit,wang2025pixnerd,ma2026frequencybooster,he2026hyperdit} have introduced architectural frequency decoupling or pixel-wise refinement modules. Specifically, DeCo \cite{ma2025deco} leverages an additional pixel decoder to generate high-frequency details conditioned on the main DiT. Similarly, DiP \cite{chen2025dip} appends a co-trained Patch Detailer Head to the global backbone to restore fine-grained local textures. PixelDiT \cite{yu2025pixeldit} adopts a dual-level architecture, utilizing a dedicated pixel-level DiT specifically for texture refinement. Unlike these approaches, our proposed PixelU achieves frequency decoupling through architectural simplicity. By integrating the $x$-prediction paradigm with simple skip connections and spatial down-sampling, PixelU entirely eliminates the need for heavyweight decoders. This design significantly reduces computational overhead while maintaining superior generation quality, offering an efficient paradigm for pixel-space diffusion.

\section{Methodology}
\label{sec:method}

\subsection{Preliminaries: Pixel Diffusion with $x$-prediction}

Flow Matching (FM) \cite{lipman2022flow} learns a velocity field that maps a prior distribution to the data distribution. Let $x$ be a clean image sampled from the data distribution and $\epsilon \sim \mathcal{N}(0, I)$ be standard Gaussian noise. We can define a noisy state $z_t$ at a continuous timestep $t \in [0, 1]$ using a linear interpolation schedule \cite{esser2024scaling,liu2022flow}:
\begin{equation}
    z_t = t x + (1 - t)\epsilon.
\end{equation}

A neural network $\operatorname{net}_{\theta}$ can be trained to predict various targets, such as the noise ($\epsilon$), velocity ($v$), or clean image ($x$). JiT \cite{li2025jit} reveals that directly predicting the clean image ($x$-prediction) offers superior performance for high-dimensional pixel-space generation compared to $\epsilon$- or $v$-prediction.

While the model learns to reconstruct $x$, the training objective is still grounded in the flow matching paradigm. The true velocity is given by the time derivative of the interpolation, yielding $v = x - \epsilon$. To compute the flow matching loss, the network's output $x_\theta$ is converted to a predicted velocity $v_\theta$:
\begin{equation}
    v_\theta = \frac{x_\theta - z_t}{1 - t},
\end{equation}
The optimization objective minimizes the mean squared error between these velocities, which implicitly acts as a dynamically reweighted $x$-prediction loss:
\begin{equation}
\label{eq:fm_loss}
    \mathcal{L}_{\text{FM}} = \mathbb{E}_{t,x,\epsilon} \left\| v_\theta - v \right\|^2 = \mathbb{E}_{t,x,\epsilon} \left\| \frac{x_\theta - x}{1 - t} \right\|^2.
\end{equation}

This design effectively benefits from both the optimization stability of $x$-prediction and the favorable sampling advantages of flow matching.

\subsection{Rethinking Decoders in Pixel Diffusion}   
\label{sec:motivation}



Concurrent modeling of low-frequency semantics and high-frequency details in the high-dimensional pixel space poses a fundamental challenge for standard DiT backbones. To tackle this challenge, recent works such as DeCo \cite{ma2025deco} and PixelDiT \cite{yu2025pixeldit} rely on pixel decoders to retain high-frequency signals. Unfortunately, these auxiliary modules introduce substantial computational burdens, significantly compromising both training and inference efficiency. In contrast, JiT \cite{li2025jit} demonstrates that remarkable efficacy can be achieved on a plain DiT simply by shifting the prediction objective. This compelling contrast leads us to rethink: \textit{Are such complex pixel decoders truly necessary?}

\begin{table}[h]
\centering
\caption{Performance comparison of different decoders and prediction targets on ImageNet 256$\times$256. $^*$ denotes our reproduced results as the official code is not publicly available. Compared to complex decoders, simple skip connections achieve the best FID under $x$-prediction without significantly increasing GFLOPs.}
\label{tab:decoder_analysis}
\small 
\setlength{\tabcolsep}{8pt} 
\resizebox{0.7\textwidth}{!}{ 
    \begin{tabular}{lccc}
    \toprule
    \multicolumn{4}{l}{\textbf{ImageNet 256$\times$256, 160 epochs, cfg=1}} \\ \midrule
    \multirow{2}{*}{\textbf{Model}} & \multirow{2}{*}{\textbf{GFLOPs}} & \multicolumn{2}{c}{\textbf{FID}} \\ \cmidrule(lr){3-4} 
    & & \textbf{$x$-pred} & \textbf{$v$-pred} \\ \midrule
    Baseline (JiT-B/16) & 21.99 & 43.65 & 190.11 \\
    + DDT decoder \cite{wang2025ddt} & 25.35 & 43.48 & 189.50  \\
    + DeCo decoder \cite{ma2025deco} & 25.91 & 41.20 & \textbf{42.07} \\
    + PixelDiT decoder$^*$ \cite{yu2025pixeldit} & 33.78 & 39.20 & 44.85 \\ \midrule
    \textbf{+ Skip Connections (Ours)} & 23.20 & \textbf{38.53} & 185.54 \\ \bottomrule
    \end{tabular}
}
\end{table}

To investigate the actual gains of these pixel decoders under different prediction targets, we conduct a series of controlled experiments on the ImageNet 256$\times$256 dataset. Building upon the recent effective pixel diffusion baseline JiT \cite{li2025jit}, we evaluate the impact of various decoder architectures on generation quality (FID and IS) under both velocity prediction ($v$-prediction) and data prediction ($x$-prediction) objectives. Observing the empirical results in Tab. \ref{tab:decoder_analysis}, we draw two key conclusions:
\begin{itemize}
    \item \textbf{Pixel decoders primarily serve $v$-prediction:} The vanilla JiT performs exceptionally poorly under $v$-prediction (FID 190.11). However, the introduction of DeCo or PixelDiT decoders yields a dramatic performance leap (dropping to 42.07 and 44.85, respectively). This indicates that such heavy decoders mainly serve to alleviate the optimization difficulties caused by $v$-prediction in the presence of high-frequency noise. 
    \item \textbf{Pixel decoders offer limited improvements for $x$-prediction:} Conversely, when employing the $x$-prediction paradigm, which anchors the generative target on clean images, the baseline model already demonstrates stable performance (FID 43.65). In this context, introducing DeCo or PixelDiT decoders at the cost of substantial GFLOPs brings only marginal performance gains (dropping to merely 41.20 and 39.20).
\end{itemize}

\subsection{Simplicity over Complexity: Skip Connections} 
Under the $x$-prediction paradigm, the network directly targets the clean image rather than the complex high-dimensional noise of $v$-prediction, making heavyweight pixel decoders essentially redundant. However, a critical challenge remains: the inherent high-frequency signals of clean images (e.g., fine textures and sharp edges) are still prone to being washed out during deep feature extraction. Other studies \cite{chen2023vision,park2022how,si2022inception} also show that transformers tend to capture low-frequency semantics well but struggle with high-frequency signals. Based on these findings, we raise a fundamental question: \textit{under the $x$-prediction paradigm, can we bypass heavyweight decoders and utilize computationally efficient operations to preserve and recover these high-frequency details?} To this end, we discard all auxiliary decoder modules and adopt a classic, computationally lightweight design—\textbf{Skip Connections}. As shown in the last row of Tab. \ref{tab:decoder_analysis}, merely introducing skip connections between the encoder and decoder allows the model to achieve an FID of 38.53 under $x$-prediction, outperforming all variants equipped with complex decoders while introducing negligible computational overhead. Since $x$-prediction explicitly requires the network to output clean pixels, skip connections provide an ``information highway'' that directly routes pristine high-frequency spatial details from shallow layers to deep layers, perfectly compensating for the spatial information loss in deep features. Notably, this simple design completely fails under $v$-prediction (FID 185.54), further corroborating that the synergy between $x$-prediction and skip connections is the key to our breakthrough in pixel diffusion.

\begin{figure}[t]
    \centering
    \begin{minipage}{0.48\textwidth}
        \centering
        \includegraphics[width=\linewidth]{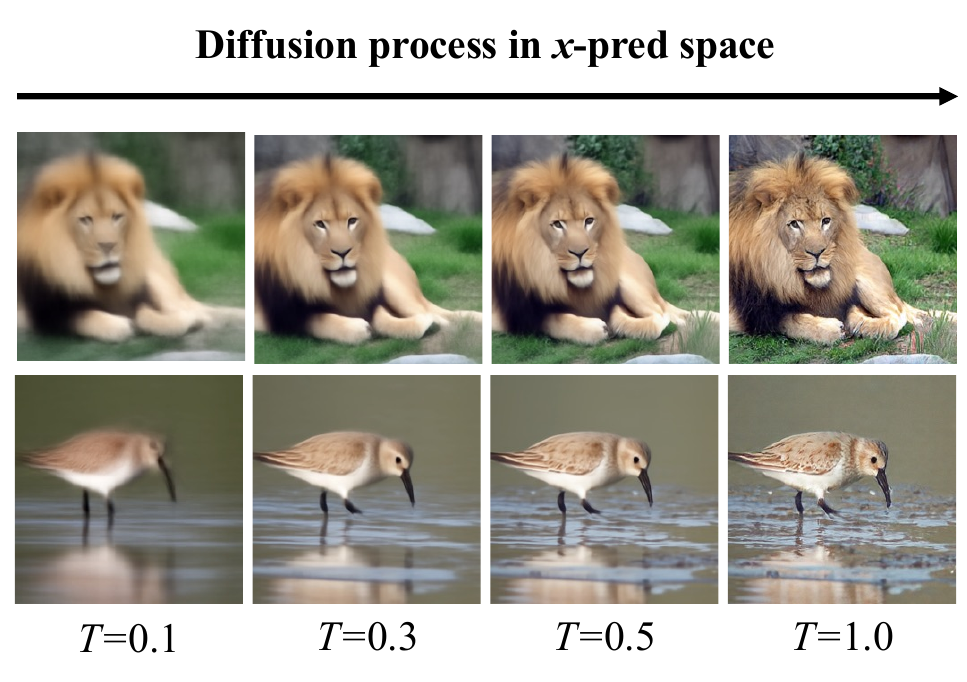} 
        \caption{Visualization of the diffusion process in $x$-prediction space via intermediate $\hat{x}_0$ at various timesteps $T$. Diffusion model establishes macroscopic low-frequency structures (e.g., basic shapes and colors) in the early stages, before progressively refining high-frequency microscopic details.}
        \label{fig:diffsuion_process}
    \end{minipage}
    \hfill 
    \begin{minipage}{0.48\textwidth}
        \centering
        \includegraphics[width=\linewidth]{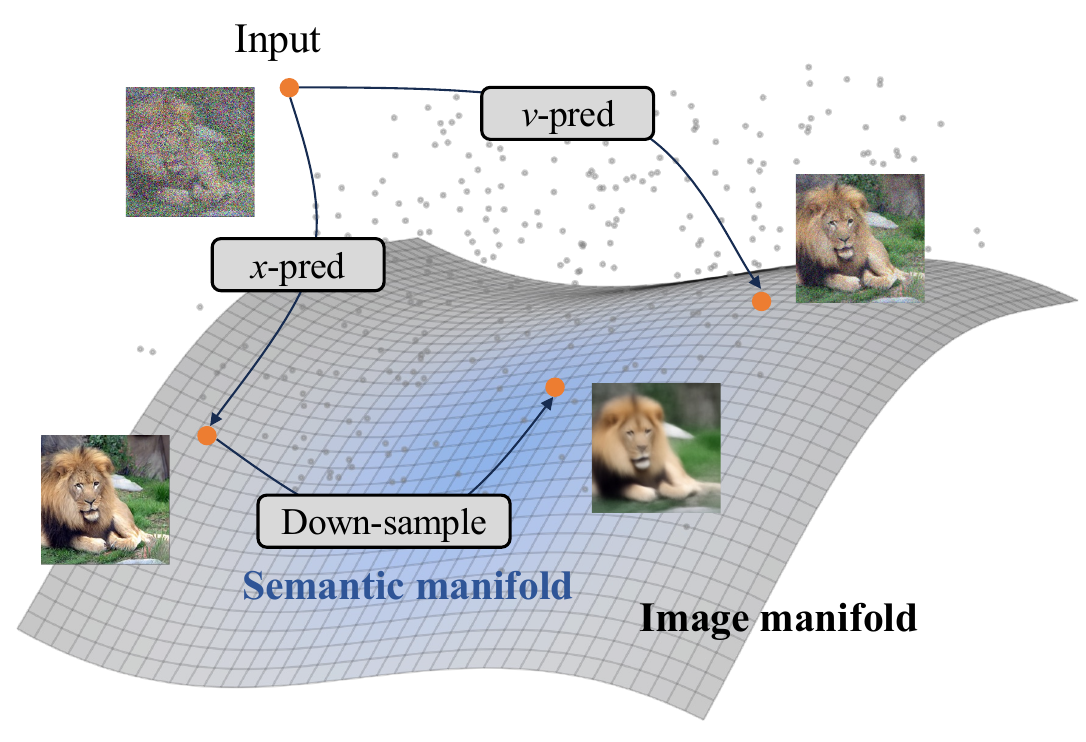} 
        \caption{Unlike $v$-prediction which points toward the high-dimensional noise space, $x$-prediction maps the noisy input directly onto the low-dimensional image manifold. Spatial down-sampling further explicitly compresses this representation further into an endogenous, low-frequency semantic manifold (highlighted in blue).}
        \label{fig:manifold}
    \end{minipage}
\end{figure}

\subsection{Down-Sampling for Low-Frequency Semantics}
\label{sec:Down-Sampling}

By delegating high-frequency details to skip connections, the backbone network is liberated to focus exclusively on modeling low-frequency global semantics. To further reveal the generative process of diffusion models, we visualize the intermediate $\hat{x}_0$ at various timesteps across the $x$-prediction trajectory. As shown in Fig. \ref{fig:diffsuion_process}, the generative process intrinsically prioritizes macroscopic, low-frequency structures during its early stages (e.g., $T=0.1$), followed by a progressive refinement of microscopic high-frequency details. The overall layout and semantic composition of the image are largely established during these initial phases.

While previous work \cite{li2025jit} reveals that natural images inherently reside on a low-dimensional manifold, our generative trajectory analysis implies the existence of a fundamental sub-space within it. As illustrated in Fig. \ref{fig:manifold}, this broader image manifold contains a highly compact, endogenous semantic sub-manifold (highlighted in blue) that is predominantly governed by low-frequency signals.

To explicitly compel the backbone network to capture this endogenous semantic manifold, we incorporate \textbf{Spatial Down-sampling} modules within the network. Inherently functioning as a natural low-pass filter, down-sampling systematically attenuates high-frequency noise and compresses the entangled features. As depicted by the mapping trajectory in Fig. \ref{fig:manifold}, this operation effectively anchors the representation from the broader image manifold directly into the compact, semantically rich sub-manifold. This aligns with existing findings \cite{williams2023unified} that down-sampling aids diffusion denoisers by automatically discarding higher-frequency subspaces dominated by noise.


\begin{figure*}[t]
    \centering
    \includegraphics[width=1.0\textwidth]{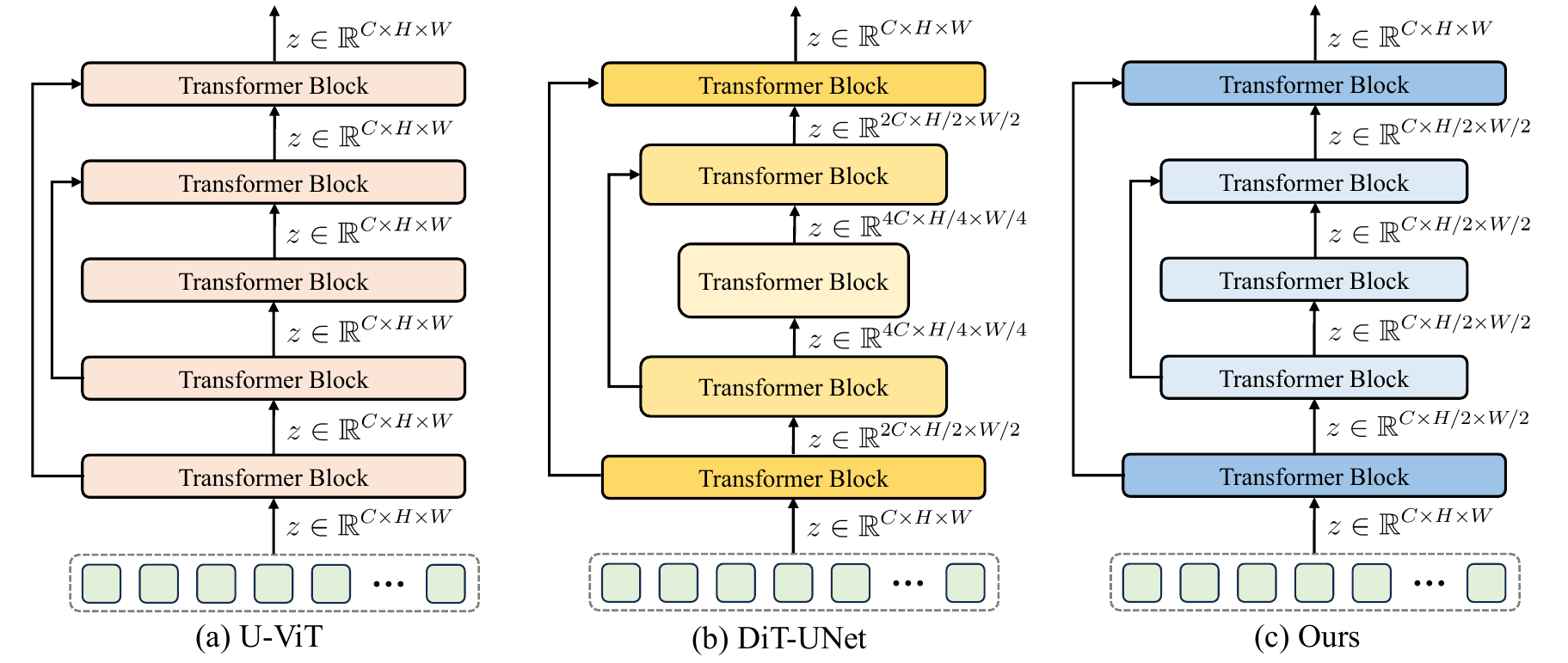}
    \caption{Architectural comparison of Diffusion Transformers. (a) U-ViT maintains a flat, full-resolution sequence ($H \times W$). (b) DiT-UNet uses multi-stage down-sampling with expanding channel dimensions (e.g., $2C, 4C$). (c) PixelU employs a minimalist design with a single stage of down-sampling ($H/2 \times W/2$) and strictly maintains a constant channel dimension throughout the network to minimize computational overhead.}
    \label{fig:main}
\end{figure*}

\subsection{PixelU Architecture}
\label{sec:pixelu}

Based on the preceding insight, we introduce \textbf{PixelU}, a minimalist U-shaped Diffusion Transformer designed specifically for pixel-space generation by integrating skip connections and spatial down-sampling. As illustrated in Fig. \ref{fig:main}, while PixelU shares a macroscopic U-shaped topology with previous models \cite{dhariwal2021diffusion,ho2020denoising,song2020denoising,song2021score}, its internal dimensional routing diverges significantly. First, unlike U-ViT \cite{bao2023uvit} (Fig. \ref{fig:main}a), which maintains a computationally expensive flat, full-resolution sequence ($H \times W$) and lacks an explicit low-frequency bottleneck, PixelU incorporates spatial down-sampling. This operation explicitly filters high-frequency noise and compresses the spatial resolution ($H/2 \times W/2$), forcing the deep layers to focus on the endogenous semantic manifold. Second, rather than adopting the traditional U-Net paradigm \cite{tian2024udit,hoogeboom2023simple} (Fig. \ref{fig:main}b) that employs multi-stage down-sampling coupled with continuous channel expansion (e.g., $2C$, $4C$), PixelU employs a drastically simplified design. It performs only a \textit{single stage} of down/up-sampling and strictly maintains a \textit{constant channel dimension} ($C$) across all intermediate layers. By avoiding multi-stage complexities and channel expansion, this streamlined design significantly mitigates the computational burden, yielding substantial reductions in GFLOPs. Ultimately, this architecture establishes a minimalist and efficient paradigm for end-to-end pixel diffusion.

\begin{figure*}[t]
    \centering
    \includegraphics[width=1.0\textwidth]{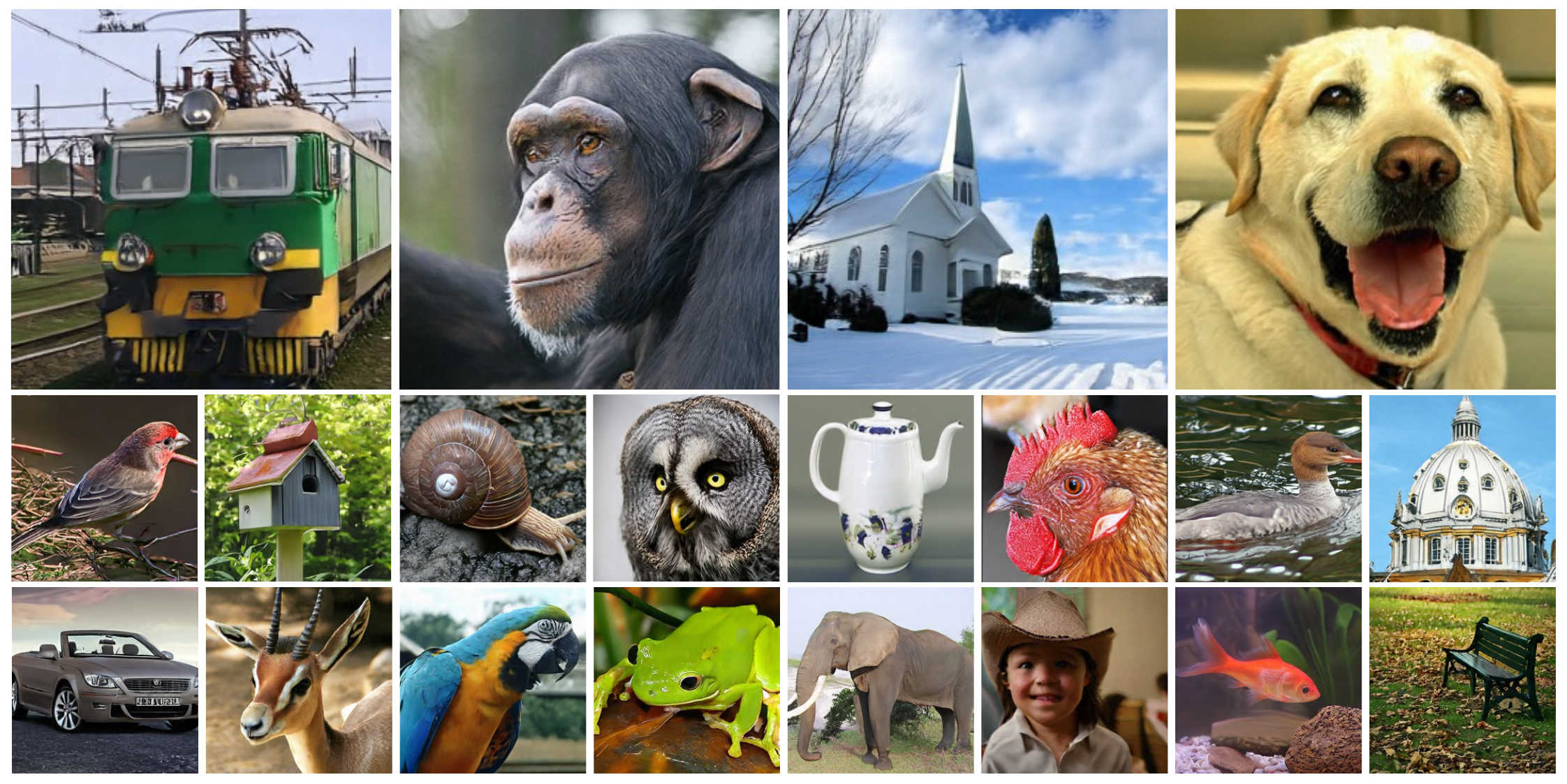}
    \caption{Selected $256 \times 256$ and $512 \times 512$ resolution samples from PixelU-H. We use a classifier-free guidance scale $\text{cfg} = 4.0$.}
    \label{fig:visual}
\end{figure*}

\section{Experiments}

\subsection{Experiment Setup}

\noindent \textbf{Model Settings.} 
To empirically validate the efficacy and scalability of PixelU, we conduct extensive experiments on ImageNet using three different model scales: Base (B), Large (L), and Huge (H). The detailed configurations for each variant are detailed in Tab. \ref{tab:configurations}.

\begin{table}[t]
    \centering
    \begin{minipage}[t]{0.52\textwidth}
        \centering
        \caption{Configurations of PixelU architecture at different model sizes.} 
        \label{tab:configurations}
        \setlength{\tabcolsep}{3pt} 
        \resizebox{\linewidth}{!}{ 
            \begin{tabular}{lccccc}
            \toprule
            \textbf{Model} & \textbf{Params} &  \textbf{Channel} & \textbf{Heads} & \textbf{Depth} \\
            \midrule
            PixelU-B/16  & 163.7M   & 768  & 12 & [4,4,4]    \\
            PixelU-B/32  & 165.8M   & 768  & 12 & [4,4,4]    \\
            \midrule
            PixelU-L/16  & 522.8M   & 1024 & 16 & [8,8,8]   \\
            PixelU-L/32  & 525.5M   & 1024 & 16 & [8,8,8]   \\
            \midrule
            PixelU-H/16 & 1168.3M   & 1152 & 16 & [8,20,8] \\
            PixelU-H/32 & 1171.8M   & 1152 & 16 & [8,20,8] \\
            \bottomrule
            \end{tabular}
        }
    \end{minipage}\hfill 
    \begin{minipage}[t]{0.44\textwidth}
        \centering
        \caption{Comparing PixelU against JiT on ImageNet 256×256 with CFG.} 
        \label{tab:baseline_compare}
        \setlength{\tabcolsep}{5pt} 
        \resizebox{\linewidth}{!}{ 
            \begin{tabular}{lccc}
            \toprule
            \textbf{Model} & \textbf{GFLOPs} & \textbf{FID$\downarrow$} & \textbf{IS$\uparrow$} \\
            \midrule
             JiT-B/16 & 25 & 3.66 & 275.1 \\
            \rowcolor[gray]{0.9} PixelU-B/16 & 21.8 & 3.40 & 276.01 \\
            \midrule
             JiT-L/16 & 88 & 2.36 & 298.5 \\
            \rowcolor[gray]{0.9} PixelU-L/16 & 73.1 & 2.17 & 295.88 \\
            \midrule
             JiT-H/16 & 182 & 1.86 & 303.4 \\
             JiT-G/16 & 383 & 1.82 & 292.6 \\
            \rowcolor[gray]{0.9} PixelU-H/16 & 136.9 & 1.63 & 305.88 \\
            \bottomrule
            \end{tabular}
        }
    \end{minipage}
    
\end{table}

\noindent \textbf{Implementation Details.}
For class-to-image generation, our implementation follows the training setup of \cite{li2025jit} on the ImageNet-1K dataset \cite{russakovsky2015imagenet} at $256 \times 256$ and $512 \times 512$ resolutions. We use a global batch size of 1024 and the AdamW optimizer with $\beta = (0.9, 0.95)$ and a constant learning rate of $2 \times 10^{-4}$ on 8$\times$B200 GPUs. Throughout the diffusion training, we employ logit-normal sampling, where $\text{logit}(t) \sim \mathcal{N}(-0.8, 0.8^2)$, and Exponential Moving Average (EMA) with a decay of 0.9999. We employ REPA \cite{yu2025repa,singh2025irepa} with a loss weight of $\lambda_{\text{repa}} = 0.01$ and apply the alignment at the middle block of the network. For evaluation, we report FID, sFID, Inception Score, and Precision–Recall on 50K samples following ADM \cite{dhariwal2021diffusion}. We use 50 Heun sampling steps with Classifier-Free Guidance (CFG) \cite{ho2022classifier} and the CFG interval \cite{kynkaanniemi2024applying}. 

\begin{table*}[t] 
\centering 
\small 
\caption{Quantitative results on ImageNet 256x256 and 512x512 with CFG. PixelU achieves superior performance in end-to-end pixel diffusion and is competitive with two-stage latent diffusion models.}
\label{tab:quantitative}
\setlength{\tabcolsep}{5pt} 

\resizebox{\textwidth}{!}{ 
\begin{tabular}{c|l|c c c c| c c c c c}
\toprule
\multicolumn{1}{c}{} & & \textbf{Params} & \textbf{Epochs} & \textbf{NFE} & \textbf{GFLOPs} & \textbf{FID}$\downarrow$ & \textbf{sFID}$\downarrow$ & \textbf{IS}$\uparrow$ & \textbf{Pre.}$\uparrow$ & \textbf{Rec.}$\uparrow$ \\
\midrule

\rowcolor{gray!20} \cellcolor{white} \multirow{29}{*}{\rotatebox[origin=c]{90}{{256$\times$256}}} & \multicolumn{10}{l}{\textit{Latent-space Diffusion}} \\ 
& \textcolor{gray}{LDM-4-G \cite{rombach2022high}} & \textcolor{gray}{400M} & \textcolor{gray}{170} & \textcolor{gray}{-} & \textcolor{gray}{-} & \textcolor{gray}{3.60} & \textcolor{gray}{-} & \textcolor{gray}{247.6} & \textcolor{gray}{0.87} & \textcolor{gray}{0.48} \\
& \textcolor{gray}{DiT-XL/2 \cite{peebles2023scalable}} & \textcolor{gray}{675M + 86M} & \textcolor{gray}{1400} & \textcolor{gray}{250$\times$2} & \textcolor{gray}{119} & \textcolor{gray}{2.27} & \textcolor{gray}{4.60} & \textcolor{gray}{278.2} & \textcolor{gray}{0.83} & \textcolor{gray}{0.57} \\
& \textcolor{gray}{SiT-XL/2 \cite{ma2024sit}} & \textcolor{gray}{675M + 86M} & \textcolor{gray}{1400} & \textcolor{gray}{250$\times$2} & \textcolor{gray}{119} & \textcolor{gray}{2.06} & \textcolor{gray}{4.50} & \textcolor{gray}{284.0} & \textcolor{gray}{0.83} & \textcolor{gray}{0.59} \\
& \textcolor{gray}{MaskDiT \cite{zheng2023fast}} & \textcolor{gray}{675M} & \textcolor{gray}{1600} & \textcolor{gray}{-} & \textcolor{gray}{-} & \textcolor{gray}{2.28} & \textcolor{gray}{5.67} & \textcolor{gray}{276.5} & \textcolor{gray}{0.80} & \textcolor{gray}{0.61} \\
& \textcolor{gray}{REPA-XL/2 \cite{yu2025repa}} & \textcolor{gray}{675M + 86M} & \textcolor{gray}{800} & \textcolor{gray}{250$\times$2} & \textcolor{gray}{119} & \textcolor{gray}{1.42} & \textcolor{gray}{4.70} & \textcolor{gray}{305.7} & \textcolor{gray}{0.80} & \textcolor{gray}{0.64} \\ 
& \textcolor{gray}{LightningDiT \cite{yao2025reconstruction}} & \textcolor{gray}{675M} & \textcolor{gray}{800} & \textcolor{gray}{-} & \textcolor{gray}{119} & \textcolor{gray}{1.35} & \textcolor{gray}{4.15} & \textcolor{gray}{295.3} & \textcolor{gray}{0.79} & \textcolor{gray}{0.65} \\
& \textcolor{gray}{SVG-XL \cite{shi2025latent}} & \textcolor{gray}{675M} & \textcolor{gray}{1400} & \textcolor{gray}{-} & \textcolor{gray}{-} & \textcolor{gray}{1.92} & \textcolor{gray}{-} & \textcolor{gray}{264.9} & \textcolor{gray}{-} & \textcolor{gray}{-} \\
& \textcolor{gray}{DDT-XL \cite{wang2025ddt}} & \textcolor{gray}{675M} & \textcolor{gray}{400} & \textcolor{gray}{-} & \textcolor{gray}{119} & \textcolor{gray}{1.26} & \textcolor{gray}{-} & \textcolor{gray}{310.6} & \textcolor{gray}{0.79} & \textcolor{gray}{0.65} \\
& \textcolor{gray}{RAE-XL \cite{zheng2025rae}} & \textcolor{gray}{839M} & \textcolor{gray}{800} & \textcolor{gray}{-} & \textcolor{gray}{146} & \textcolor{gray}{1.13} & \textcolor{gray}{-} & \textcolor{gray}{262.6} & \textcolor{gray}{0.78} & \textcolor{gray}{0.67} \\
\rowcolor{gray!20} \cellcolor{white} & \multicolumn{10}{l}{\textit{Pixel-space Diffusion}} \\ 
& StyleGAN-XL \cite{sauer2022stylegan} & - & - & - & - & 2.30 & 4.02 & 265.1 & 0.78 & 0.53 \\
& ADM-G \cite{dhariwal2021diffusion} & 554M & 400 & 250 & 1120 & 4.59 & 5.25 & 186.7 & \textbf{0.82} & 0.52 \\
& RDM \cite{teng2023relay} & 553M + 553M & 400 & - & - & 1.99 & \textbf{3.99} & 260.4 & 0.81 & 0.58 \\
& CDM \cite{ho2022cascaded} & - & 2160 & - & - & 4.88 & - & 158.7 & - & - \\
& RIN \cite{jabri2022scalable} & 410M & 480 & - & 334 & 3.42 & - & 182.0 & - & - \\
& VDM++ \cite{kingma2023understanding} & 2B & - & 250$\times$2 & 555 & 2.12 & - & 267.7 & - & - \\
& JetFormer \cite{tschannen2024jetformer} & 2.8B & - & - & - & 6.64 & - & - & 0.69 & 0.56 \\
& Simple Diffusion \cite{hoogeboom2023simple} & 2B & - & 250$\times$2 & 555 & 2.44 & - & 256.3 & - & - \\
& FractalMAR-H \cite{li2025fractal} & 848M & 600 & - & - & 6.15 & - & \textbf{348.9} & 0.81 & 0.46 \\
& FARMER \cite{zheng2025farmer} & 1.9B & 320 & - & - & 3.60 & - & 269.2 & 0.81 & 0.51 \\
& EPG \cite{lei2025epg} & 583M & 800 & - & - & 2.04 & - & 283.2 & 0.80 & 0.56 \\
& PixelFlow-XL/4 \cite{chen2025pixelflow} & 677M & 320 & 120$\times$2 & 2909 & 1.98 & 5.83 & 282.1 & 0.81 & 0.60 \\
& PixNerd-XL/16 \cite{wang2025pixnerd} & 700M & 320 & {100$\times$2} & 134 & 1.95 & 4.54 & 300 & 0.80 & 0.60 \\
& JiT-G \cite{li2025jit} & 2B & 600 & 100$\times$2 & 383 & 1.82 & - & 292.6 & 0.79 & 0.62 \\
& DeCo-XL/16 \cite{ma2025deco} & 682M & 600 & 100$\times$2 & - & 1.69 & 4.59 & 304 & 0.79 & 0.63 \\
& PixelGen-XL/16 \cite{ma2026pixelgen} & 676M & 160 & 100$\times$2 & - & 1.83 & 4.59 & 293.6 & 0.79 & 0.63 \\
\cline{2-11}
& PixelU-H/16 & 1168M & 320 & 100$\times$2 & 137 & 1.77 & 5.29 & 306.86 & 0.78 & 0.63 \\
& PixelU-H/16 & 1168M & 600 & 100$\times$2 & 137 & \textbf{1.63} & 5.04 & 305.88 & 0.79 & \textbf{0.64} \\
\midrule

\rowcolor{gray!20} \cellcolor{white} \multirow{12}{*}{\rotatebox[origin=c]{90}{{512$\times$512}}} & \multicolumn{10}{l}{\textit{Latent-space Diffusion}} \\ 
& \textcolor{gray}{DiT-XL/2 \cite{peebles2023scalable}} & \textcolor{gray}{675M + 86M} & \textcolor{gray}{600} & \textcolor{gray}{250$\times$2} & \textcolor{gray}{525} & \textcolor{gray}{3.04} & \textcolor{gray}{5.02} & \textcolor{gray}{240.8} & \textcolor{gray}{0.84} & \textcolor{gray}{0.54} \\
& \textcolor{gray}{SiT-XL/2 \cite{ma2024sit}} & \textcolor{gray}{675M + 86M} & \textcolor{gray}{600} & \textcolor{gray}{250$\times$2} & \textcolor{gray}{525} & \textcolor{gray}{2.62} & \textcolor{gray}{4.18} & \textcolor{gray}{252.2} & \textcolor{gray}{0.84} & \textcolor{gray}{0.57} \\ 
\rowcolor{gray!20} \cellcolor{white} & \multicolumn{10}{l}{\textit{Pixel-space Diffusion}} \\ 
& ADM-G \cite{dhariwal2021diffusion} & 554M & 400 & 250 & 1983 & 7.72 & 6.57 & 172.7 & \textbf{0.87} & 0.53 \\
& RIN \cite{jabri2022scalable} & 320M & - & 250 & 415 & 3.95 & - & 216.0 & - & - \\
& SimpleDiffusion \cite{hoogeboom2023simple} & 2B & 800 & 250$\times$2 & 555 & 3.54 & - & 205.0 & - & - \\
& VDM++ \cite{kingma2023understanding} & 2B & 800 & 250$\times$2 & 555 & 2.65 & - & 278.1 & - & - \\
& PixNerd-XL/16 \cite{wang2025pixnerd} & 700M & 340 & 100$\times$2 & 583 & 2.84 & 5.95 & 245.6 & 0.80 & 0.59 \\
& JiT-H/32 \cite{li2025jit} & 682M & 600 & 100$\times$2 & 183 & 1.94 & - & 309.1 & - & - \\
& DeCo-XL/16 \cite{ma2025deco} & 682M & 340 & 100$\times$2 & - & 2.22 & \textbf{4.67} & 290.0 & 0.80 & \textbf{0.60} \\
\cline{2-11}
& PixelU-H/32 & 1152M & 600 & 100$\times$2 & 138 & \textbf{1.92} & 5.98 & \textbf{322.10} & 0.80 & 0.58 \\
\bottomrule
\end{tabular}
}
\end{table*}

\subsection{Class-to-Image Generation}

\noindent \textbf{ImageNet 256$\times$256.} 
As summarized in Tab. \ref{tab:quantitative} and Tab. \ref{tab:baseline_compare}, our PixelU-H/16 achieves an exceptional FID of 1.63 and an IS of 305.88 after 600 training epochs, surpassing recent state-of-the-art pixel-space models. Notably, PixelU consistently outperforms the JiT \cite{li2025jit} baseline across various model sizes while maintaining a substantially lower computational cost. This efficiency advantage scales remarkably to larger models: our PixelU-H/16 drastically outperforms the massive JiT-G (FID 1.82) while utilizing significantly fewer parameters (1.15B vs. 2B) and demanding only about 1/3 of the computational cost (137 GFLOPs vs. 383 GFLOPs). It also surpasses competitive architectures like DeCo-XL/16 \cite{ma2025deco} (FID 1.69) and completely eclipses earlier models such as ADM \cite{dhariwal2021diffusion} and PixelFlow-XL/4 \cite{chen2025pixelflow}. Furthermore, PixelU demonstrates remarkable convergence efficiency: at just 320 training epochs, it already reaches an FID of 1.76, decisively beating PixNerd-XL/16 \cite{wang2025pixnerd} (1.95) and PixelFlow-XL/4 (1.98) trained for similar durations. We note that while the concurrent PixelDiT \cite{yu2025pixeldit} and SiD2 \cite{hoogeboom2025simpler} reports a FID of 1.61 and 1.38, its implementation and training pipelines remain closed-source. For fairness, we include its FID for reference but exclude it from Tab. \ref{tab:quantitative}. When compared to the latent-space paradigm, our end-to-end PixelU outperforms widely adopted baselines such as DiT-XL/2 \cite{peebles2023scalable} (FID 2.27) and SiT-XL/2 \cite{ma2024sit} (FID 2.06) using less than half the training epochs (600 vs. 1400 epochs). While advanced latent models equipped with pre-trained autoencoders (e.g., DDT \cite{wang2025ddt}, RAE \cite{zheng2025rae}) still hold slightly lower FIDs, PixelU significantly narrows this gap without relying on lossy VAE compressions, proving the immense potential of pure pixel-space modeling.

\noindent \textbf{ImageNet 512$\times$512.} 
Scaling end-to-end pixel diffusion to higher resolutions presents significant challenges due to the quadratic increase in spatial complexity. Despite this, at the 512$\times$512 resolution, our PixelU-H/16 achieves a leading FID of 1.92 and an IS of 322 after 600 epochs. PixelU consistently outperforms existing pixel-based baselines, including DeCo-XL/16 \cite{ma2025deco} (FID 2.22) and PixNerd-XL/16 \cite{wang2025pixnerd} (FID 2.84), as well as larger 2B-parameter models like VDM++ \cite{kingma2023understanding} (FID 2.65) and SimpleDiffusion \cite{hoogeboom2023simple} (FID 3.54). Furthermore, PixelU demonstrates clear advantages over representative two-stage latent diffusion models. Compared to DiT-XL/2 \cite{peebles2023scalable} (FID 3.04) and SiT-XL/2 \cite{ma2024sit} (FID 2.62) trained for 600 epochs, PixelU delivers significantly better image fidelity. 

\subsection{Analysis}
\begin{figure*}[t]
    \centering
    \includegraphics[width=1.0\textwidth]{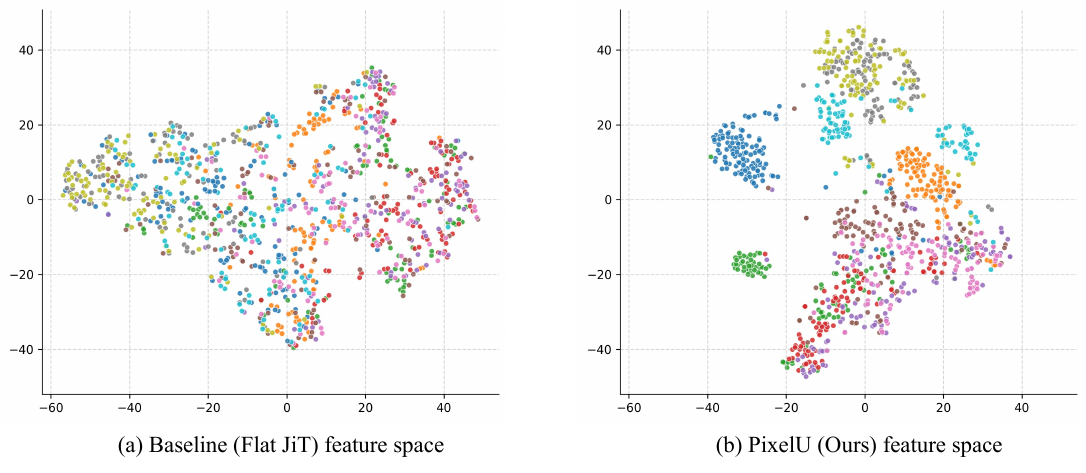}
    \caption{Feature visualization with t-SNE for 10 ImageNet classes (100 random samples per class), with each class shown in a distinct color. These features are extracted from the intermediate blocks of both the baseline (JiT) and PixelU.}
    \label{fig:tsne}
\end{figure*}

\noindent \textbf{Feature visualization with t-SNE.}
To further validate the efficacy of our architectural design in capturing low-frequency semantics, we visualize the intermediate feature spaces of the baseline (Flat JiT) and our PixelU using t-SNE. As illustrated in Fig. \ref{fig:tsne}(a), the feature representations of the flat baseline are scattered and heavily entangled across different classes. This occurs because the full-resolution architecture lacks a structural bottleneck, leaving its feature manifold highly susceptible to interference from high-frequency textures and irrelevant pixel-level noise. Conversely, Fig. \ref{fig:tsne}(b) demonstrates that PixelU generates remarkably compact and well-separated class clusters. This striking contrast visually confirms our hypothesis: the spatial down-sampling mechanism successfully acts as an explicit low-pass filter. By systematically shedding redundant, class-irrelevant high-frequency noise, it compels the bottleneck layers to learn a highly pure and discriminative global semantic representation.

\begin{figure*}[t]
    \centering
    \includegraphics[width=1.0\textwidth]{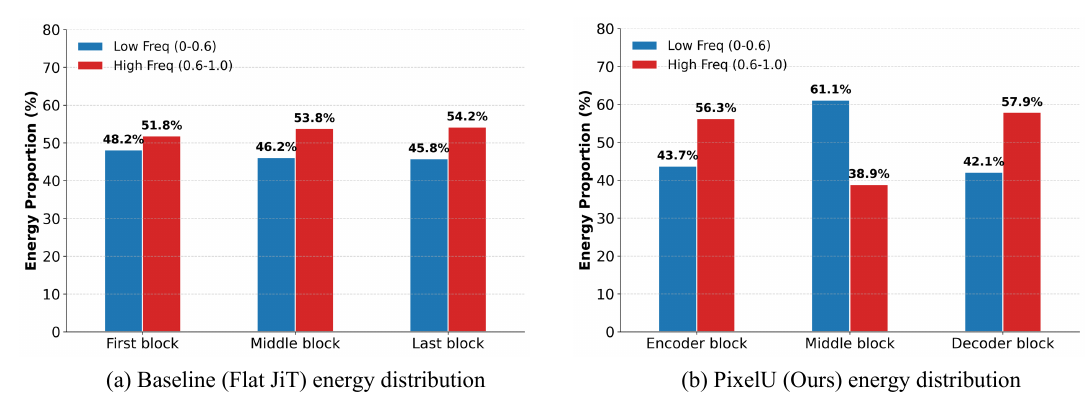}
    \caption{\textbf{Frequency energy distribution across different blocks.} We normalize the radial frequency to $r \in [0, 1]$ and calculate the energy proportions within the low-frequency ($r \le 0.6$) and high-frequency ($r > 0.6$) bands.}
    \label{fig:freq_comparison}
\end{figure*}

\noindent \textbf{Frequency Energy Distribution.}
To further validate the superiority of our frequency decoupling mechanism, we present a comparative analysis of the frequency energy distribution between the baseline (flat JiT) and our proposed PixelU (as shown in Figure \ref{fig:freq_comparison}). Specifically, we apply a 2D Fast Fourier Transform (2D-FFT) to the spatial feature maps of different blocks to compute their channel-averaged magnitude spectra. The flat architecture of the baseline JiT suffers from persistent frequency entanglement, where high-frequency noise and low-frequency semantics remain tightly coupled across all layers. Lacking an explicit spatial bottleneck, this isotropic design forces the backbone to model both simultaneously, significantly exacerbating optimization difficulty in pixel space. Conversely, PixelU demonstrates an explicit and highly structured frequency dynamic. While the shallow encoder captures abundant high-frequency spatial details, the middle bottleneck exhibits a precipitous drop in high-frequency energy. This confirms that our spatial down-sampling acts as an endogenous low-pass filter, effectively isolating a pure, low-dimensional semantic manifold. Subsequently, high-frequency energy sharply rebounds in the decoder block, proving that skip connections function as an information highway, seamlessly routing pristine high-frequency details to the output to compensate for the bottleneck's compression. Collectively, these quantitative observations substantiate that PixelU achieves elegant frequency decoupling through the highly efficient synergy of spatial down-sampling and skip connections.

\subsection{Ablation Study}
\noindent \textbf{Contribution of Core Components.}
Table \ref{tab:ablation_components} presents an ablation study on the core components of our PixelU architecture: spatial down-sampling and skip connections. Starting from the full-resolution baseline (FID 43.65, 21.99 GFLOPs), integrating only skip connections improves generation quality (FID 38.53) by directly propagating spatial details across layers. Conversely, applying spatial down-sampling in isolation degrades the generative performance (FID 48.87) despite reducing the computational cost to 18.25 GFLOPs. This degradation occurs because the down-sampling bottleneck inevitably discards high-frequency details. However, combining both components achieves the optimal performance (FID 34.92) while maintaining a lower computational overhead (19.46 GFLOPs) than the baseline. This demonstrates their crucial synergy: down-sampling efficiently extracts low-frequency semantics and reduces computational costs, while skip connections effectively compensate by recovering the filtered high-frequency details. Together, they form a highly efficient paradigm for end-to-end pixel-space generation. 

\begin{table}[t]
    \centering
    \caption{\textbf{Ablation study on core components.} The combination of skip connections and spatial down-sampling yields the best generative performance while effectively reducing the computational cost compared to the baseline.}
    \label{tab:ablation_components}
    \setlength{\tabcolsep}{7pt} 
    \resizebox{0.8\textwidth}{!}{%
    \begin{tabular}{ll cc | cc}
        \toprule
        \multicolumn{6}{l}{\textbf{ImageNet 256$\times$256, 160 epochs, cfg=1}} \\ 
        \midrule
        \multirow{2}{*}{\textbf{GFLOPs}} & \multirow{2}{*}{\textbf{Params}} & \multicolumn{2}{c|}{\textbf{Metrics}} & \multirow{2}{*}{\begin{tabular}[c]{@{}c@{}}\textbf{Skip} \\ \textbf{Connections}\end{tabular}} & \multirow{2}{*}{\textbf{Downsampling}} \\
        \cmidrule{3-4}
        & & \textbf{FID}$\downarrow$ & \textbf{IS} $\uparrow$ & & \\
        \midrule
        21.99 & 131.1M     & 43.65 & 34.08 & \xmark   & \xmark   \\
        18.25 & 153.6M     & 48.87 & 30.29 & \xmark & \cmark   \\
        23.20 & 135.8M     & 38.53 & 38.57 & \cmark   & \xmark \\
        19.46 & 158.3M & \textbf{34.92} & \textbf{43.51} & \cmark   & \cmark   \\
        \bottomrule
    \end{tabular}%
    } 
\end{table}

\begin{table}[t]
    \centering
    \newcommand{\colA}{2.0cm} 
    \newcommand{\colB}{1.6cm} 
    \newcommand{\colC}{1.4cm} 
    \newcommand{\colD}{1.4cm} 

    \begin{minipage}{0.48\textwidth}
        \centering
        \caption{Ablation study of the block allocation strategy.}
        \label{tab:depth}
        \resizebox{\linewidth}{!}{
            \begin{tabular}{wc{\colA} wc{\colB} wc{\colC} wc{\colD}}
                \toprule
                \multicolumn{4}{l}{\textbf{ImageNet 256$\times$256, 160 epochs, cfg=1}} \\ 
                \midrule
                \textbf{Depth} & \textbf{GFLOPs} & \textbf{FID}$\downarrow$ & \textbf{IS}$\uparrow$ \\ 
                \midrule
                $[5,2,5]$ & 22.48 & 35.47 & 41.56 \\ 
                \rowcolor[gray]{0.9} $[4,4,4]$ & 19.46 & \textbf{34.92} & \textbf{43.51} \\ 
                $[3,6,3]$ & 16.44 & 38.08 & 40.25 \\ 
                \bottomrule
            \end{tabular}
        }
    \end{minipage}\hfill
    \begin{minipage}{0.48\textwidth}
        \centering
        \caption{Ablation study of the number of down-sampling (DS) stages.}
        \label{tab:downsample}
        \resizebox{\linewidth}{!}{
            \begin{tabular}{wc{\colA} wc{\colB} wc{\colC} wc{\colD}}
                \toprule
                \multicolumn{4}{l}{\textbf{ImageNet 256$\times$256, 160 epochs, cfg=1}} \\ 
                \midrule
                \textbf{Depth} & \textbf{DS Stages} & \textbf{FID}$\downarrow$ & \textbf{IS}$\uparrow$ \\ 
                \midrule
                $12$ & 0 & 38.53 & 38.57 \\ 
                \rowcolor[gray]{0.9} $[4,4,4]$ & 1 & \textbf{34.92} & \textbf{43.51} \\ 
                $[2,2,4,2,2]$ & 2 & 42.70 & 36.40 \\ 
                \bottomrule
            \end{tabular}
        }
    \end{minipage}
\end{table}

\noindent \textbf{Block Allocation Strategy.} Tab. \ref{tab:depth} shows that a balanced $[4, 4, 4]$ configuration achieves the optimal FID of 34.92. Allocating insufficient blocks to the middle bottleneck (e.g., $[5, 2, 5]$) restricts the network's capacity to explore the low-frequency semantic manifold. Conversely, reducing the encoder/decoder blocks (e.g., $[3, 6, 3]$) compromises high-frequency detail extraction and fusion via skip connections. Thus, a balanced allocation best harmonizes high-frequency and low-frequency modeling.

\noindent \textbf{Number of Down-sampling Stages.} Tab. \ref{tab:downsample} validates our single-stage down-sampling design. A flat architecture (0 stages) struggles with frequency entanglement, yielding a suboptimal FID of 38.53. Conversely, employing 2 down-sampling stages degrades the FID to 42.70. Because PixelU strictly maintains a constant channel dimension, multiple spatial compressions create an overly restrictive bottleneck that irreversibly destroys crucial representations. Consequently, a single down-sampling stage optimally decouples frequencies without catastrophic information loss.

\noindent \textbf{General-purpose Improvements.} 
To demonstrate the compatibility of our architecture with standard diffusion enhancements, we progressively integrate several established techniques into the vanilla PixelU baseline. As shown in Tab. \ref{tab:improvements}, incorporating in-context class tokens \cite{li2025jit,li2024autoregressive} and REPA \cite{yu2025repa,singh2025irepa} yields steady, incremental gains in both FID and IS across different model scales. Subsequently, the application of Classifier-Free Guidance (CFG) \cite{ho2022classifier} alongside the CFG interval \cite{kynkaanniemi2024applying} provides a substantial leap in generative quality, drastically reducing the FID of PixelU-H/16 from 6.95 to 2.42 at 160 epochs. Finally, extending the training schedule to 600 epochs allows the network to fully converge, achieving our optimal performance of 1.63 FID for the PixelU-H/16 model.

\begin{table}[t]
    \centering
    \caption{\textbf{Ablation study of general-purpose improvements.} Progressive integration of established techniques into the vanilla PixelU baseline consistently enhances generation quality across different model scales.}
    \label{tab:improvements}
    \setlength{\tabcolsep}{4pt} 
    \resizebox{0.9\textwidth}{!}{
        \begin{tabular}{l r c c c c c c} 
            \toprule
            \multirow{2}{*}{\textbf{Model Components}} & \multirow{2}{*}{\textbf{Epoch}} & \multicolumn{3}{c}{\textbf{PixelU-B/16}} & \multicolumn{3}{c}{\textbf{PixelU-H/16}} \\
            \cmidrule(r){3-5} \cmidrule(l){6-8}
            & & \textbf{GFLOPs} & \textbf{FID}$\downarrow$ & \textbf{IS}$\uparrow$ & \textbf{GFLOPs} & \textbf{FID}$\downarrow$ & \textbf{IS}$\uparrow$ \\
            \midrule
            PixelU (vanilla) & 160 & 19.46 & 34.92 & 43.51 & 117.86 & 12.41 & 95.37 \\
            \midrule
            + In-context class tokens & 160 & 21.42 & 30.88 & 51.84 & 136.32 & 8.93 & 124.06 \\
            + REPA & 160 & 21.76 & 26.86 & 58.46 & 136.88 & 6.95 & 142.78 \\
            \midrule
            \multirow{2}{*}{+ CFG and CFG interval} & 160 & 21.76 & 4.39 & 230.41 & 136.88 & 2.42 & 290.26 \\
            & 600 & 21.76 & \textbf{3.40} & \textbf{276.01} & 136.88 & \textbf{1.63} & \textbf{305.88} \\
            \bottomrule
        \end{tabular}%
    }
\end{table}


\section{Conclusion}

In this work, we present PixelU, a minimalist, single-stage U-shaped Diffusion Transformer tailored for efficient pixel-space generation. We demonstrate that under the $x$-prediction paradigm, complex pixel decoders used in prior works become largely redundant. Instead, PixelU advocates for simplicity by leveraging zero-cost skip connections to directly route uncorrupted high-frequency spatial details from shallow to deep layers. Concurrently, a constant-channel spatial down-sampling mechanism acts as a natural low-pass filter, allowing the backbone to focus exclusively on a compact, low-frequency semantic manifold. Extensive experiments show that this elegant frequency decoupling enables PixelU to surpass recent end-to-end pixel-space models at a significantly lower computational cost. While latent diffusion models currently still retain a slight edge, PixelU significantly narrows this gap without the lossy decoding artifacts inherent in autoencoders. Ultimately, our findings highlight that fundamental architectural designs can also establish a powerful and highly efficient new paradigm for end-to-end pixel diffusion.

\clearpage
\appendix

\setcounter{figure}{0}
\setcounter{table}{0}
\setcounter{equation}{0}
\renewcommand\thesection{\Alph{section}}
\renewcommand\thefigure{A\arabic{figure}}
\renewcommand\thetable{A\arabic{table}}
\renewcommand\theequation{A\arabic{equation}}


\section{More Implementary Details}

\subsection{Model Configuration}

Tab. \ref{tab:hyperparameters} summarizes the experiment configurations for PixelU-B, PixelU-L, and PixelU-H. In practice, Our implementation closely follows the public codebases such as DiT \cite{peebles2023scalable}, SiT \cite{ma2024sit} and JiT \cite{li2025jit}. The source code is provided in the \textit{.zip} file.

\begin{table}[h]
\centering
\small 
\setlength{\tabcolsep}{1pt} 
\newcolumntype{C}{>{\centering\arraybackslash}p{2.0cm}} 
\caption{\textbf{Hyperparameter settings across different model scales.}}
\label{tab:hyperparameters}

\begin{tabular}{l|CCC}
 & \textbf{PixelU-B} & \textbf{PixelU-L} & \textbf{PixelU-H} \\
\hline
\rowcolor{gray!20} \multicolumn{4}{l}{\textbf{Architecture}} \\
depth & [4,4,4] & [8,8,8] & [8,20,8] \\
hidden dim & 768 & 1024 & 1280 \\
heads & 12 & 16 & 16 \\
image size & \multicolumn{3}{c}{256 (other settings: 512)} \\
patch size & \multicolumn{3}{c}{\texttt{image\_size} / 16} \\
bottleneck & \multicolumn{3}{c}{128 (B/L), 256 (H)} \\
dropout & \multicolumn{3}{c}{0 (B/L), 0.1 (H)} \\
in-context class tokens & \multicolumn{3}{c}{32 (if used)} \\
in-context start block & 4 & 8 & 8 \\
\hline
\rowcolor{gray!20} \multicolumn{4}{l}{\textbf{REPA settings}} \\
Alignment depth & 6 & 12 & 18 \\
Loss weight $\lambda_{\text{repa}}$ & \multicolumn{3}{c}{0.01} \\
Alignment encoder & \multicolumn{3}{c}{DINOv2-B with registers \cite{darcet2023vision}} \\
\hline
\rowcolor{gray!20} \multicolumn{4}{l}{\textbf{Training}} \\
epochs & \multicolumn{3}{c}{160 (ablation), 600} \\
warmup epochs \cite{goyal2017accurate}  & \multicolumn{3}{c}{5} \\
optimizer & \multicolumn{3}{c}{Adam \cite{kingma2014adam}, $\beta_1, \beta_2 = 0.9, 0.95$} \\
batch size & \multicolumn{3}{c}{1024} \\
learning rate & \multicolumn{3}{c}{2e-4} \\
learning rate schedule & \multicolumn{3}{c}{constant} \\
weight decay & \multicolumn{3}{c}{0} \\
ema decay & \multicolumn{3}{c}{0.9999} \\
time sampler & \multicolumn{3}{c}{$\text{logit}(t) \sim \mathcal{N}(\mu, \sigma^2)$, $\mu = -0.8, \sigma = 0.8$} \\
noise scale & \multicolumn{3}{c}{$1.0 \times \texttt{image\_size} / 256$} \\
clip of $(1 - t)$ in division & \multicolumn{3}{c}{0.05} \\
class token drop (for CFG) & \multicolumn{3}{c}{0.1} \\
\hline
\rowcolor{gray!20} \multicolumn{4}{l}{\textbf{Sampling}} \\
ODE solver & \multicolumn{3}{c}{Heun [20]} \\
ODE steps & \multicolumn{3}{c}{50} \\
time steps & \multicolumn{3}{c}{linear in [0.0, 1.0]} \\
CFG scale sweep range \cite{ho2022classifier} & \multicolumn{3}{c}{[1.0, 4.0]} \\
CFG interval \cite{kynkaanniemi2024applying} & \multicolumn{3}{c}{[0.1, 1] (if used)} \\
\end{tabular}
\end{table}

\subsection{Representation Alignment}
Specifically, we use DINOv2 with registers \cite{darcet2023vision} as the representation alignment (REPA) \cite{yu2025repa,singh2025irepa} model. We interpolate the input images to $224 \times 224$ and set patch size to 14, yielding 256 tokens which is consistent with the sequence length of the diffusion model. 

In our PixelU architecture, applying REPA at the centering block introduces a spatial resolution mismatch between the down-sampled PixelU bottleneck features and ViT encoder features. To resolve this issue, we follow U-REPA \cite{tian2025urepa} by upscaling the smaller-sized PixelU features rather than downscaling the target ViT features. Specifically, the lower-resolution feature from the PixelU middle block is first passed through an MLP projector and subsequently upsampled to match the exact spatial dimensions of the ViT features. This operation order effectively bridges the architectural disparity while maintaining an optimal balance between semantic fidelity and computational efficiency.

\subsection{In-context class conditioning}

Following JiT \cite{li2025jit}, we employ in-context class conditioning by prepending class tokens to the feature sequence. However, we empirically observe that injecting these tokens at the initial layer degrades the model's generative performance. Consistent with the findings in \cite{li2025jit}, we mitigate this issue by introducing the class tokens starting from the middle stage of our PixelU architecture.

During the spatial up-sampling phase in the decoder, we first decouple the class tokens from the image tokens. The up-sampling operation (pixel un-shuffling) is then applied exclusively to the image tokens to restore their spatial resolution. Finally, the unaltered class tokens are concatenated back with the up-sampled image tokens before passing them to the subsequent transformer blocks.

\section{Frequency-Domain Analysis of Prediction Targets}

In this section, we mathematically formalize why $v$-prediction requires heavyweight decoders to handle high-frequency (HF) components in pixel space, whereas $x$-prediction naturally mitigates this. Furthermore, we reveal why this fundamental flaw is masked in Latent Diffusion Models (LDMs).

\subsection{The High-Frequency Bottleneck in Pixel Space}

Let $x \in \mathbb{R}^D$ be a clean image drawn from the data distribution $p_{data}$, and $\epsilon \sim \mathcal{N}(0, I)$ be the standard Gaussian noise. By design, $x$ and $\epsilon$ are statistically independent. Under a linear schedule, the noisy state at timestep $t$ is $z_t = t x + (1-t)\epsilon$, and the target velocity for $v$-prediction is defined as $v = x - \epsilon$.

We analyze these signals in the frequency domain using the continuous Fourier transform operator $\mathcal{F}$. The frequency-domain representation of the $v$-prediction target is:
\begin{equation}
    \mathcal{F}(v)(\omega) = \mathcal{F}(x)(\omega) - \mathcal{F}(\epsilon)(\omega),
\end{equation}
where $\omega$ represents the spatial frequency. Assuming zero-mean high-frequency components, the variance of the Fourier coefficients at a specific frequency $\omega$ is characterized by the Power Spectral Density (PSD), denoted as $S(\omega) = \mathbb{E}[|\mathcal{F}(\cdot)(\omega)|^2]$. 

Because the natural image $x$ and the sampled noise $\epsilon$ are statistically independent, their cross-power spectral density is zero. Therefore, the variance of the $v$-prediction target decomposes linearly:
\begin{equation}
    S_v(\omega) = S_x(\omega) + S_\epsilon(\omega),
\end{equation}
According to the well-established spectral properties of natural images \cite{field1987relations,torralba2003statistics}, the PSD of $x$ follows a power-law decay:
\begin{equation}
    S_x(\omega) \propto \frac{1}{||\omega||^\alpha},
\end{equation}
where typically $\alpha \approx 2$. This implies that natural images possess concentrated energy in low frequencies, while high-frequency energy rapidly vanishes:
\begin{equation}
    \lim_{||\omega|| \to \infty} S_x(\omega) = 0,
\end{equation}
Conversely, the standard Gaussian noise $\epsilon$ is spatially uncorrelated, yielding a flat power spectrum across all frequencies:
\begin{equation}
    S_\epsilon(\omega) = c \quad (c > 0),
\end{equation}
Evaluating the targets at the high-frequency limit ($||\omega|| \to \infty$), we observe a fundamental divergence in optimization difficulty:
\begin{equation}
    \lim_{||\omega|| \to \infty} S_v(\omega) = \lim_{||\omega|| \to \infty} \left( S_x(\omega) + S_\epsilon(\omega) \right) = 0 + c = c
\end{equation}
\begin{equation}
    \lim_{||\omega|| \to \infty} S_x(\omega) = 0.
\end{equation}

In $v$-prediction, the target variance at high frequencies remains bounded away from zero ($S_v(\omega) \to c$). To minimize the MSE loss, the neural network is forced to explicitly memorize and regress the exact high-dimensional realizations of unstructured noise $\epsilon$. This inherently demands massive parameter capacity and complex localized modeling, heavily relying on heavyweight pixel decoders. In stark contrast, for $x$-prediction, the target variance at high frequencies naturally decays to zero ($S_x(\omega) \to 0$). The network merely needs to predict a deterministic-like absence of energy (smoothness) at high frequencies. This optimization landscape is mathematically much simpler and perfectly manageable using lightweight skip connections rather than dense parameterization.

\subsection{Prediction Targets in Latent Space}

A natural question arises: why is the catastrophic optimization difficulty of $v$-prediction seemingly absent in LDMs?

In LDMs, raw images are aggressively down-sampled (e.g., $8\times$) into a compact latent space using a pre-trained Variational Autoencoder (VAE) \cite{rombach2022high}. When an image is reduced to such a small grid, it physically loses the capacity to store rapid spatial changes (i.e., high-frequency details like fine textures and sharp edges). Therefore, the VAE effectively acts as a severe low-pass filter that permanently removes these high-frequency spatial signals.

As noted in previous method \cite{li2025jit}, \textit{when a low-dimensional space (e.g., image latent) is used, the difficulty of predicting noise is alleviated, yet at the same time is hidden rather than solved. The VAE masks the fundamental flaw of $v$-prediction by discarding the high-frequency spectrum entirely before diffusion even begins.} Because the diffusion process in LDMs operates entirely within this compressed, low-frequency subspace, the network never has to deal with the complex, high-dimensional noise $\epsilon$ that exposes the inherent capacity bottleneck of $v$-prediction. However, in end-to-end pixel space diffusion, the network must directly process all spatial details down to the raw pixel level. This exposes the high-frequency optimization problem, making the shift to $x$-prediction critical.

\section{Additional Ablation Studies}

\subsection{Skip Connections Design}
This section investigates the impact of different feature fusion methods within the skip connections on model performance, with results shown in Table~\ref{tab:skip_compare}. We compare three common designs \cite{bao2023uvit}: direct element-wise addition ($x_m + x_s$), addition followed by a linear layer Linear($x_m + x_s$), and concatenation followed by a linear layer Linear(Concat($x_m, x_s$)). Empirical results demonstrate that concatenation followed by a linear layer most effectively preserves and fuses high-frequency details from shallow layers with low-frequency semantics from deep layers, achieving the optimal generation quality (FID 26.86, IS 58.46). Consequently, we adopt this as the default skip connection design in PixelU.

\subsection{Downsampler Design}
Table~\ref{tab:downsampler_compare} compares three downsampler architectures: Max pooling, strided convolution, and Pixel Unshuffle. To maintain a constant channel capacity, we prepend a depthwise convolution to compress channels to $1/4$ before the Pixel Unshuffle operation, which inherently quadruples them. Results show that our Pixel Unshuffle block achieves the lowest FID (26.86). Unlike max pooling, which irreversibly discards high-frequency information, or strided convolutions, the information-lossless Pixel Unshuffle spatially rearranges features into the channel dimension.  This optimally preserves fine-grained details, making it our default downsampling strategy.

\begin{table}[h]
    \centering
    \begin{minipage}[t]{0.48\textwidth}
        \centering
        \caption{Ablations on the choice of skip connections with REPA and in-context class tokens.} 
        \label{tab:skip_compare}
        \setlength{\tabcolsep}{5pt} 
        \resizebox{\linewidth}{!}{ 
        \begin{tabular}{lcc}
        \toprule
        \multicolumn{3}{c}{\textbf{ImageNet 256$\times$256, 160 epochs, cfg=1}} \\
        \midrule
        \textbf{Model} & \textbf{FID$\downarrow$} & \textbf{IS$\uparrow$} \\
        \midrule
        $x_m +x_s$  & 30.11 & 52.68 \\
        \rowcolor[gray]{0.9} Linear(Concat($x_m, x_s$)) & \textbf{26.86} & 58.46 \\
        Linear($x_m +x_s$) & 28.95 & 53.61 \\
        \bottomrule
        \end{tabular}
        }
    \end{minipage}\hfill 
    \begin{minipage}[t]{0.48\textwidth}
        \centering
        \caption{Ablations on the choice of downsampler with REPA and in-context class tokens.} 
        \label{tab:downsampler_compare}
        \setlength{\tabcolsep}{5pt} 
        \resizebox{\linewidth}{!}{ 
        \begin{tabular}{lcc}
        \toprule
        \multicolumn{3}{c}{\textbf{ImageNet 256$\times$256, 160 epochs, cfg=1}} \\
        \midrule
        \textbf{Model} & \textbf{FID$\downarrow$} & \textbf{IS$\uparrow$} \\
        \midrule
        Max pooling & 27.16 & 60.07 \\
        \rowcolor[gray]{0.9} Pixel Unshuffle & \textbf{26.86} & 58.46 \\
        2D Convolution & 27.78 & 56.62 \\
        \bottomrule
        \end{tabular}
        }
    \end{minipage}
\end{table}

\begin{table}[t]
\centering
\caption{Ablation on REPA configurations with in-context class tokens. Depth of conducting REPA loss, loss weight $\lambda_{\text{repa}}$, and projector type are included.}
\label{tab:repa}
\setlength{\tabcolsep}{5pt} 
\resizebox{0.6\linewidth}{!}{
    \begin{tabular}{ccccc}
    \toprule
    \multicolumn{5}{l}{\textbf{ImageNet 256$\times$256, 160 epochs, cfg=1}} \\
    \midrule
    \textbf{Depth} & \textbf{Weight} & \textbf{Projector Type} & \textbf{FID}$\downarrow$ & \textbf{IS}$\uparrow$ \\
    \midrule
    $-$ & $-$ & $-$ & 30.88 & 51.84 \\
    \cellcolor{gray!20} 4 & 0.01 & Conv & 27.91 & 57.11 \\
    \cellcolor{gray!20} 6 & 0.01 & Conv & \textbf{26.86} & 58.46 \\
    \cellcolor{gray!20} 8 & 0.01 & Conv & 26.91 & 59.62 \\
    \cellcolor{gray!20} 10 & 0.01 & Conv & 27.05 & 58.32 \\
    \midrule
    5 & \cellcolor{gray!20} 0.005 & Conv & 27.61 & 57.62 \\
    5 & \cellcolor{gray!20} 0.01 & Conv & \textbf{26.86} & 58.46 \\
    5 & \cellcolor{gray!20} 0.05 & Conv & 26.94 & 59.51 \\
    5 & \cellcolor{gray!20} 0.1 & Conv & 28.04 & 59.09 \\
    \midrule
    5 & 0.01 & \cellcolor{gray!20} MLP & 27.67 & 55.76 \\
    5 & 0.01 & \cellcolor{gray!20} Linear & 28.24 & 56.07 \\
    5 & 0.01 & \cellcolor{gray!20} Conv & \textbf{26.86} & 58.46 \\
    \bottomrule
    \end{tabular}
} 
\end{table}

\subsection{REPA Configurations}
To fully unleash the potential of REPA \cite{yu2025repa,singh2025irepa}, we conduct a systematic ablation study on its key hyperparameters in Table~\ref{tab:repa}, including the alignment depth (Depth), alignment loss weight ($\lambda_{\text{repa}}$), and projector type.
\begin{itemize}
    \item \textbf{Alignment Depth:} Compared to the baseline without REPA (FID 30.88), introducing REPA yields significant performance improvements. The model achieves the optimal FID (26.86) when feature alignment occurs at the middle stages of the network (Depth=6). Aligning at excessively shallow or deep layers leads to a slight performance degradation.
    \item \textbf{Loss Weight:} Fixing the depth, we examine different values for $\lambda_{\text{repa}}$. Setting the weight to $0.01$ strikes the best balance between matching the encoder's representations and the diffusion generative task.
    \item \textbf{Projector Type:} In terms of the feature projector, the convolution-based mapping layer (Conv) \cite{singh2025irepa} outperforms both the multi-layer perceptron (MLP) \cite{yu2025repa} and the simple linear layer (Linear).
\end{itemize}
Based on these experimental results, we adopt Depth=6, $\lambda_{\text{repa}}=0.01$, and the Conv projector as the optimal REPA configuration for PixelU.

\begin{table}[t]
    \centering
    \caption{Training step for PixelU.}
    \label{alg:training_step}
    \begin{tabularx}{\linewidth}{@{}X@{}}
        \toprule
        \textbf{Algorithm 1} Training step \\
        \midrule
        \begin{tabular}{@{}l@{}}
            \textcolor{myteal}{\texttt{\#} $net_\theta$\texttt{: PixelU network}} \\
            \textcolor{myteal}{\texttt{\#} $x$\texttt{: training batch}} \\
            \textcolor{myteal}{\texttt{\#} $c$\texttt{: class label or textual prompt}} \\
            \\
            $t = \textcolor{mypink}{\mathtt{sample\_t}}()$ \\
            $\epsilon = \textcolor{mypink}{\mathtt{randn\_like}}(x)$ \\
            \\
            $x_t = t \mathbin{\textcolor{myblue}{*}} x \mathbin{\textcolor{myblue}{+}} (1 \mathbin{\textcolor{myblue}{-}} t) \mathbin{\textcolor{myblue}{*}} \epsilon$ \\
            $v_t = (x \mathbin{\textcolor{myblue}{-}} x_t) \mathbin{\textcolor{myblue}{/}} (1 \mathbin{\textcolor{myblue}{-}} t)$ \\
            \\
            $x_\theta = net_\theta(x_t, t, c)$ \\
            $v_\theta = (x_\theta \mathbin{\textcolor{myblue}{-}} x_t) \mathbin{\textcolor{myblue}{/}} (1 \mathbin{\textcolor{myblue}{-}} t)$ \\
            \\
            $\mathbf{loss}_{\mathrm{FM}} = \textcolor{mypink}{\mathtt{12\_loss}}(v_\theta \mathbin{\textcolor{myblue}{-}} v_t)$ \\

            $\mathbf{loss} = \mathbf{loss}_{\mathrm{FM}} \mathbin{\textcolor{myblue}{+}} \lambda_{\text{repa}}
            \mathbf{loss}_{\mathrm{REPA}}$
        \end{tabular} \\
        \bottomrule
    \end{tabularx}

\end{table}


\begin{table}[t]
    \centering
    \caption{Sampling step for PixelU.}
    \label{alg:sampling}
    \begin{tabularx}{\linewidth}{@{}X@{}}
        \toprule
        \textbf{Algorithm 2} Sampling step \\
        \midrule
        \begin{tabular}{@{}l@{}}
            \textcolor{myteal}{\texttt{\#} $x_t$\texttt{: current samplers at} $t$} \\
            \\
            $x_\theta = net_\theta(x_t, t, c)$ \\
            $v_\theta = (x_\theta \mathbin{\textcolor{myblue}{-}} x_t) \mathbin{\textcolor{myblue}{/}} (1 \mathbin{\textcolor{myblue}{-}} t)$ \\
            \\
            $x_{t\_next} = x_t \mathbin{\textcolor{myblue}{+}} (t\_next \mathbin{\textcolor{myblue}{-}} t) \mathbin{\textcolor{myblue}{*}} v_\theta$
        \end{tabular} \\
        \bottomrule
    \end{tabularx}
\end{table}

\section{Pseudocodes for PixelU}

To provide a clear overview of the implementation, we outline the pseudocodes of PixelU in the following sections. Specifically, Algorithm \ref{alg:training_step} details the training phase, which builds upon the JiT \cite{li2025jit} pipeline and incorporates REPA loss \cite{yu2025repa,singh2025irepa} to enhance representation learning. Furthermore, the pseudocodes for the sampling steps are provided in Algorithm \ref{alg:sampling}.

\section{More Visualization Results}
We show additional qualitative results for ImageNet $256 \times 256$ and $512 \times 512$ from our top-performing model: PixelU-H, as visualized in \crefrange{appendix1}{appendix8}.

\begin{figure}[h]
    \centering
    \begin{minipage}{0.7\textwidth}
        \centering
        \includegraphics[width=\linewidth]{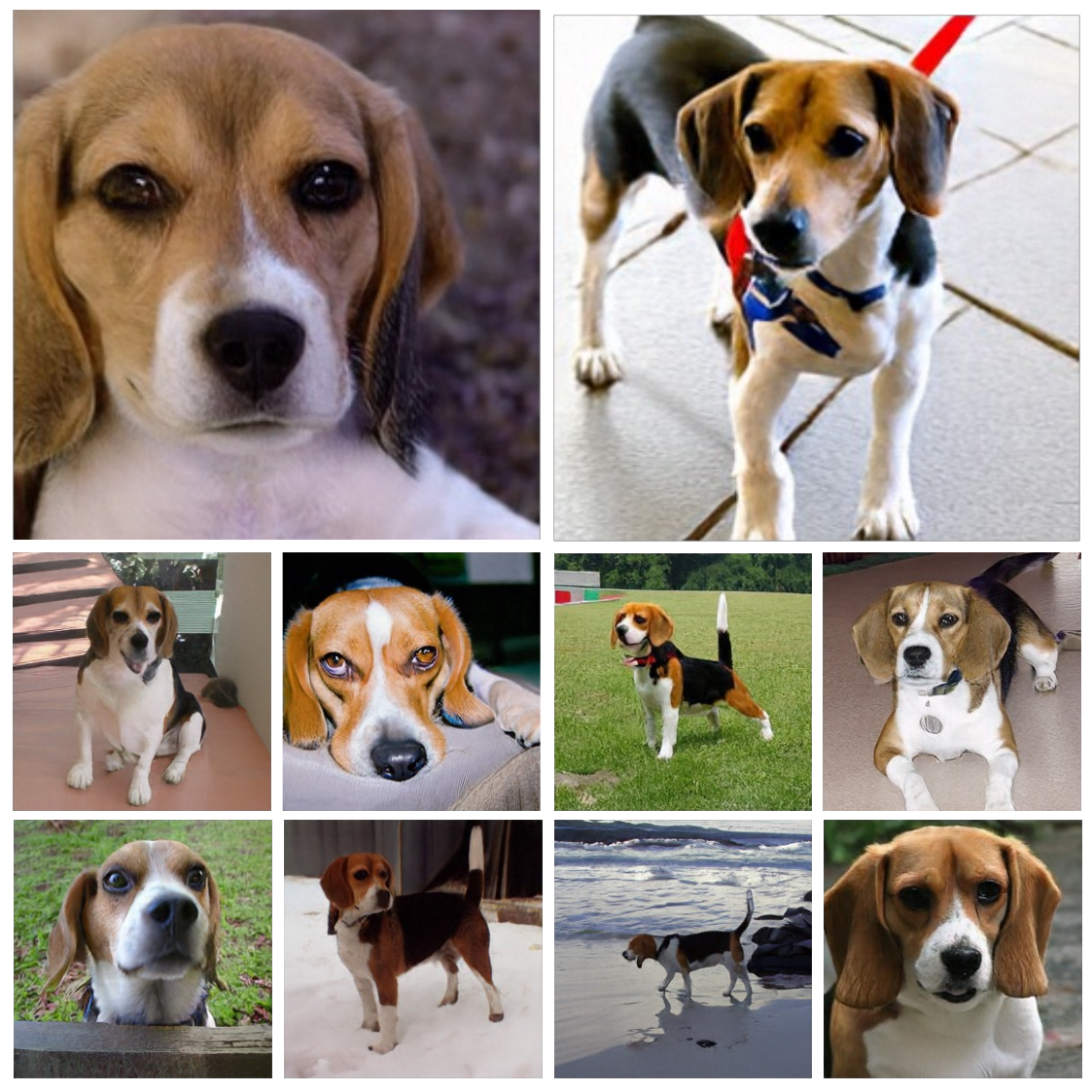} 
        \caption{Class-to-image 256×256 and 512×512 PixelU-H samples. Class 162: beagle. CFG scale = 4.0.}
        \label{appendix1}
    \end{minipage}
    \hfill 
    \begin{minipage}{0.7\textwidth}
        \centering
        \includegraphics[width=\linewidth]{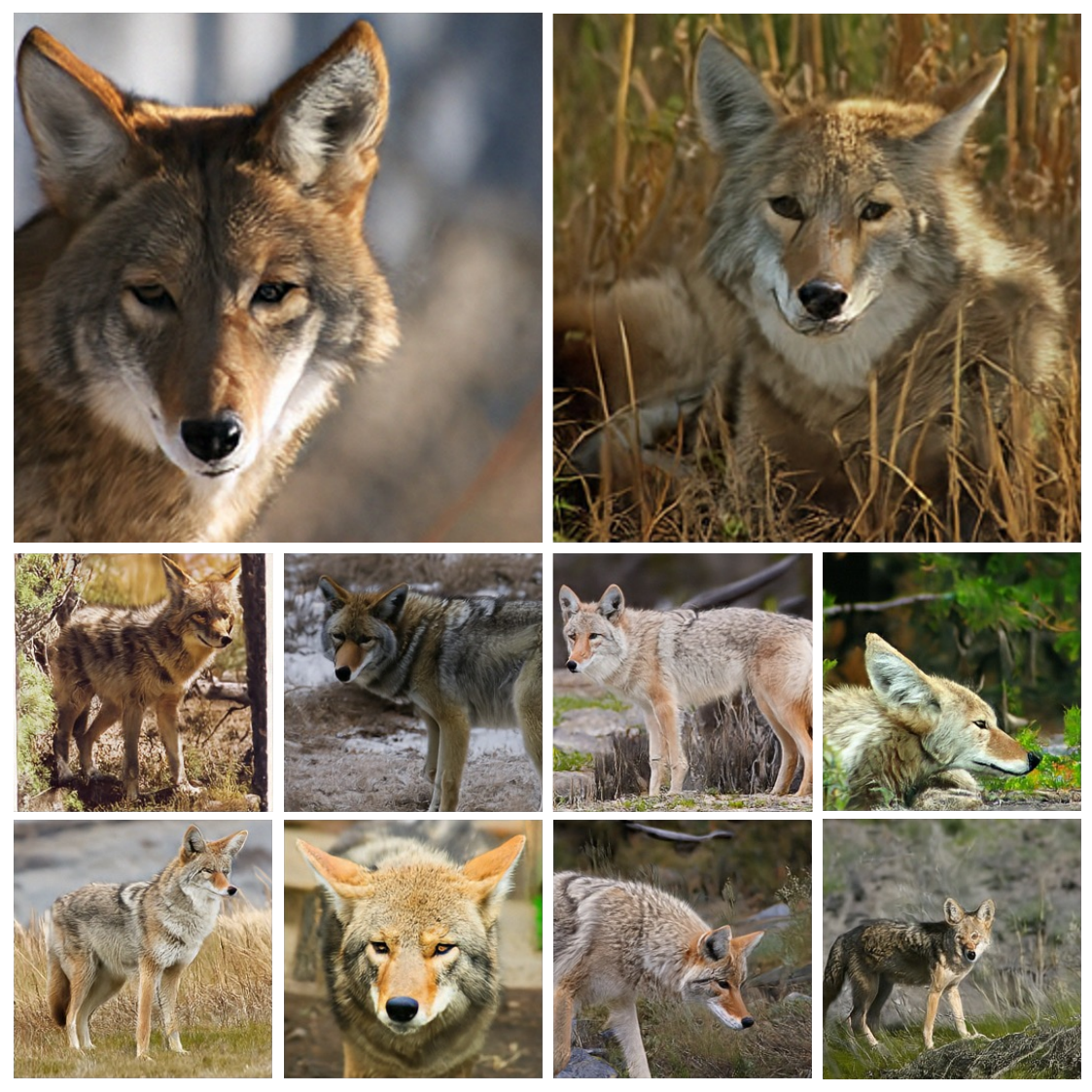} 
        \caption{Class-to-image 256×256 and 512×512 PixelU-H samples. Class 272: coyote. CFG scale = 4.0.}
        \label{appendix2}
    \end{minipage}
\end{figure}

\begin{figure}[h]

    \centering
    \begin{minipage}{0.7\textwidth}
        \centering
        \includegraphics[width=\linewidth]{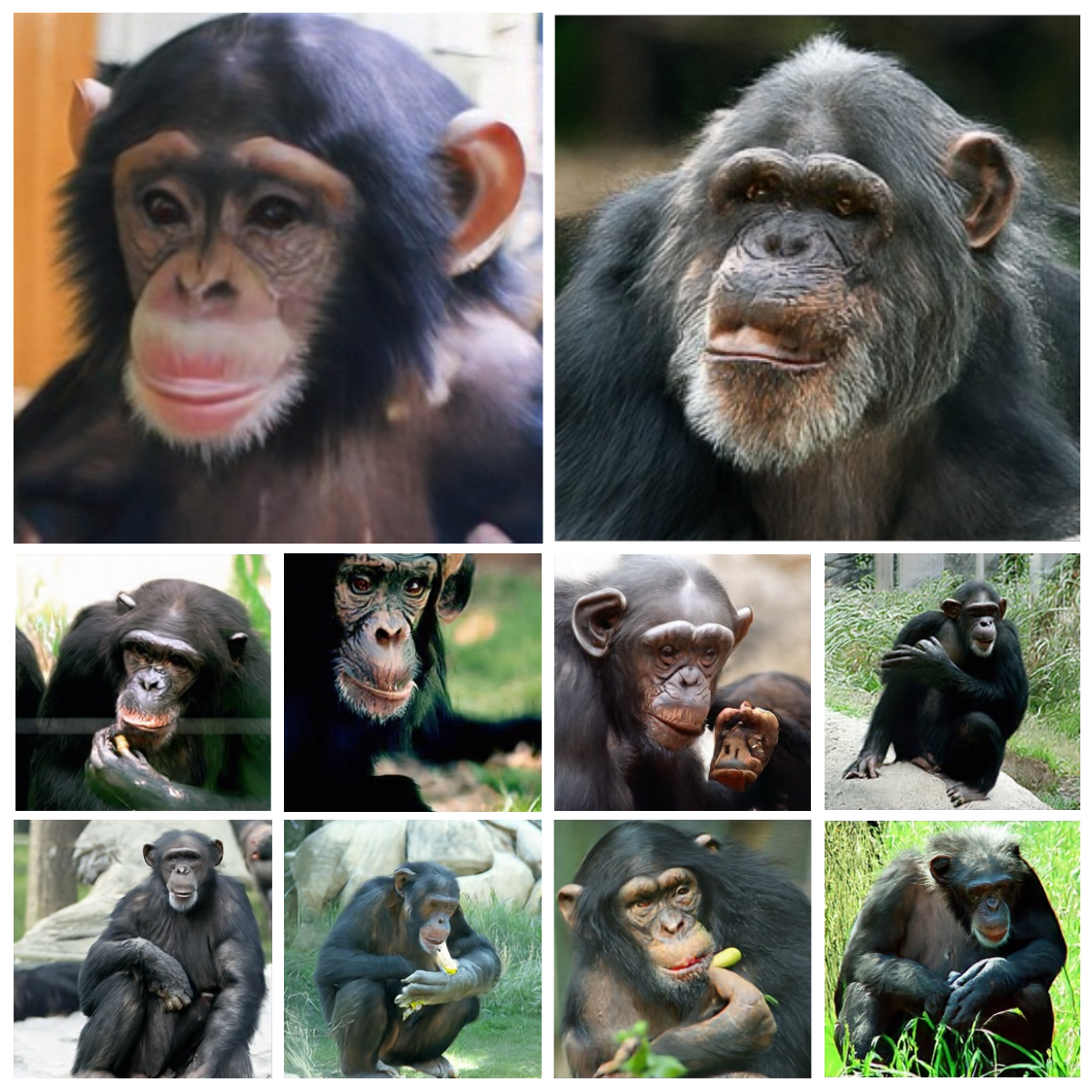} 
        \caption{Class-to-image 256×256 and 512×512 PixelU-H samples. Class 367: chimpanzee. CFG scale = 4.0.}
        \label{appendix3}
    \end{minipage}
    \hfill 
    \begin{minipage}{0.7\textwidth}
        \centering
        \includegraphics[width=\linewidth]{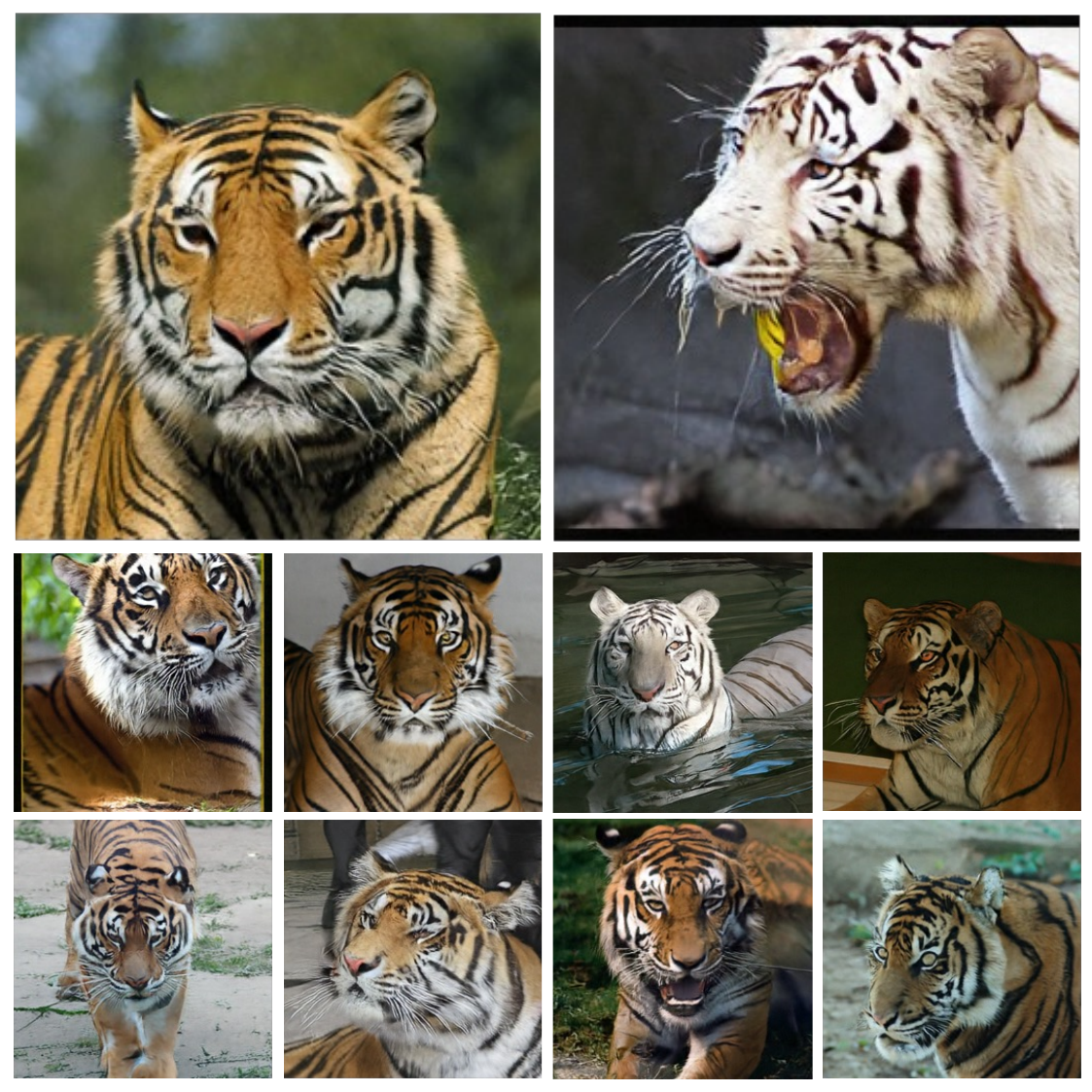} 
        \caption{Class-to-image 256×256 and 512×512 PixelU-H samples. Class 292: tiger. CFG scale = 4.0.}
        \label{appendix4}
    \end{minipage}
\end{figure}

\clearpage

\begin{figure}[h]
    \centering
    \begin{minipage}{0.7\textwidth}
        \centering
        \includegraphics[width=\linewidth]{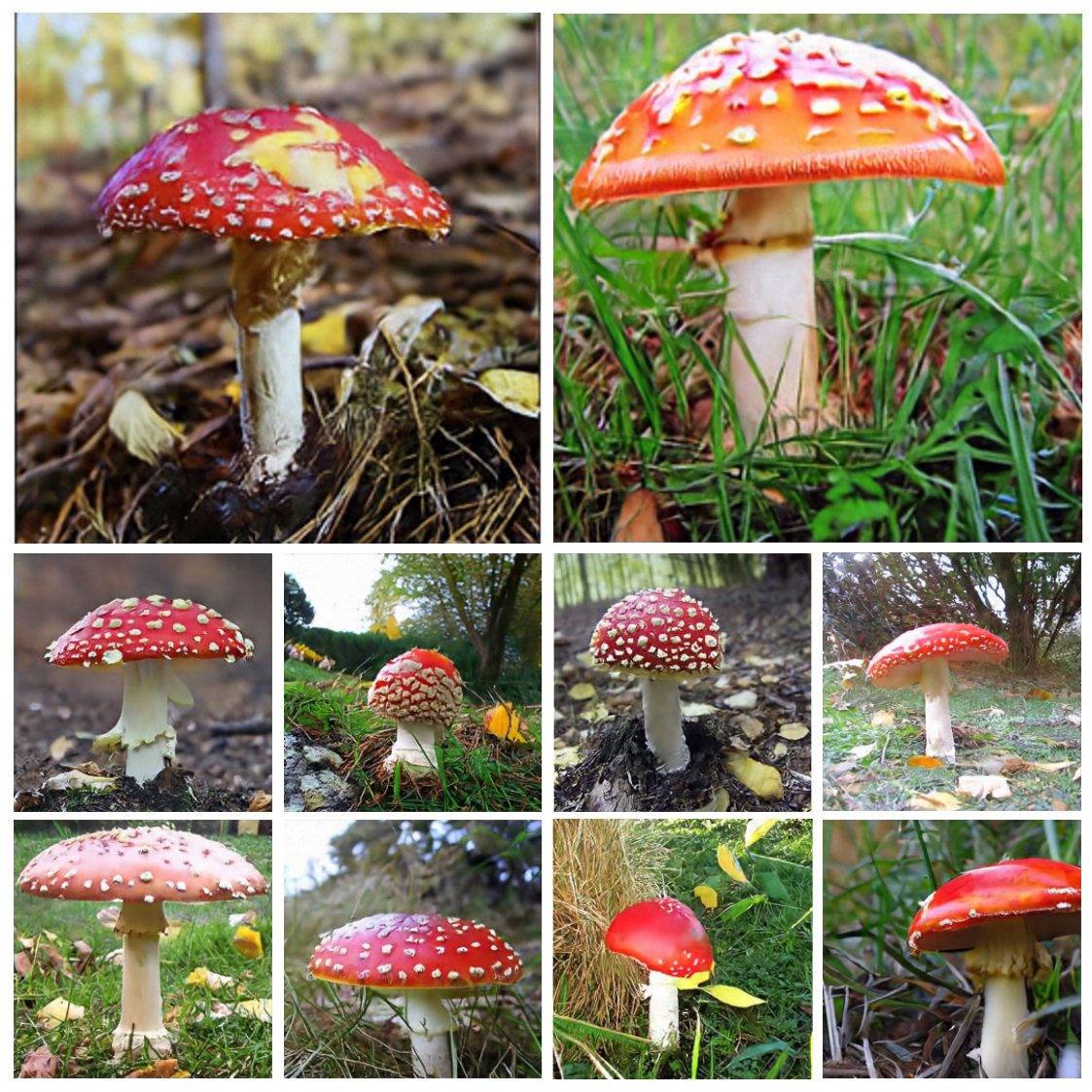} 
        \caption{Class-to-image 256×256 and 512×512 PixelU-H samples. Class 992: agaric. CFG scale = 4.0.}
        \label{appendix5}
    \end{minipage}
    \hfill 
    \begin{minipage}{0.7\textwidth}
        \centering
        \includegraphics[width=\linewidth]{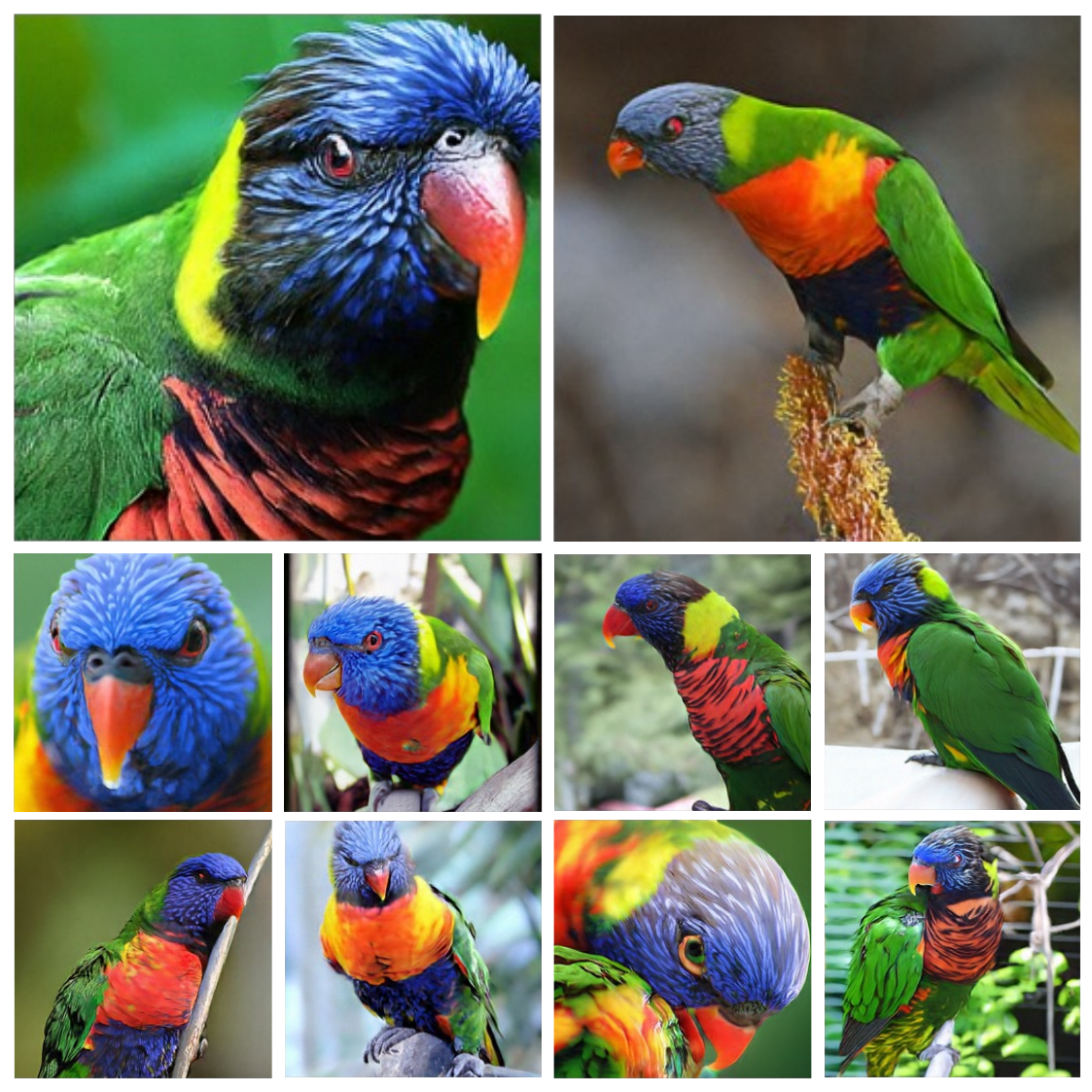} 
        \caption{Class-to-image 256×256 and 512×512 PixelU-H samples. Class 90: lorikeet. CFG scale = 4.0.}
        \label{appendix6}
    \end{minipage}
\end{figure}

\begin{figure}[h]
    \centering
    \begin{minipage}{0.7\textwidth}
        \centering
        \includegraphics[width=\linewidth]{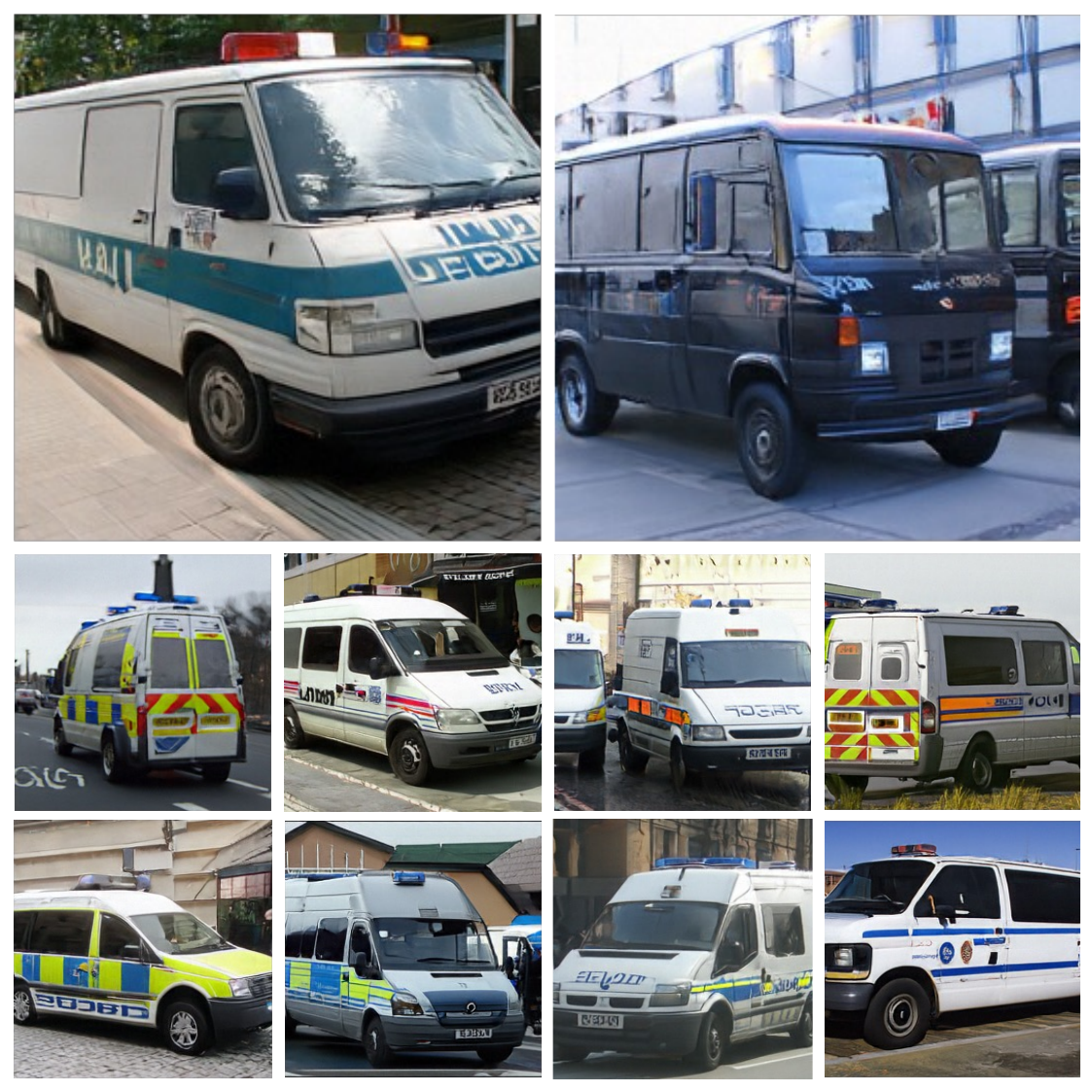} 
        \caption{Class-to-image 256×256 and 512×512 PixelU-H samples. Class 734: police van. CFG scale = 4.0.}
        \label{appendix7}
    \end{minipage}
    \hfill 
    \begin{minipage}{0.7\textwidth}
        \centering
        \includegraphics[width=\linewidth]{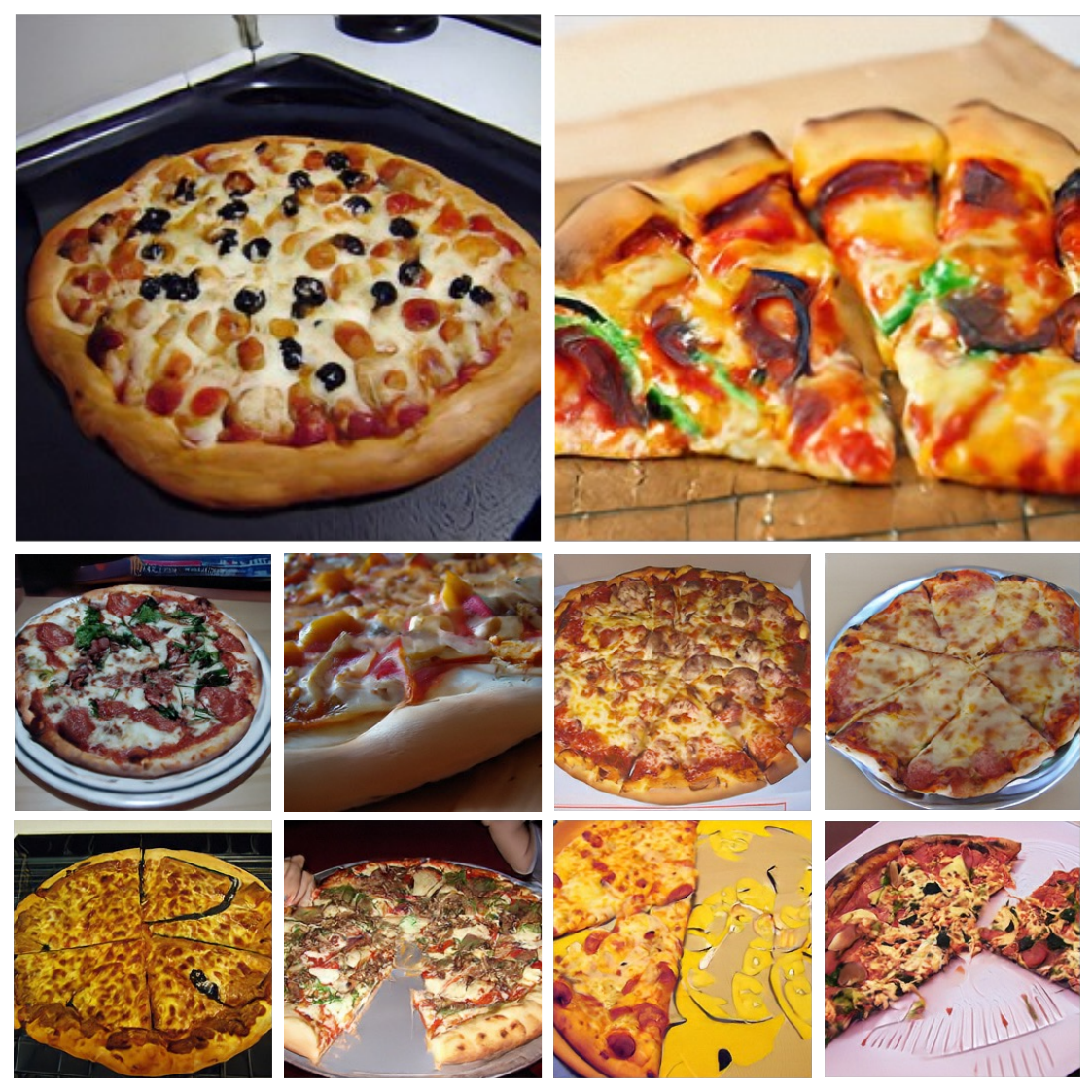} 
        \caption{Class-to-image 256×256 and 512×512 PixelU-H samples. Class 963: pizza. CFG scale = 4.0.}
        \label{appendix8}
    \end{minipage}
\end{figure}

\clearpage  

%
%
\bibliographystyle{splncs04}
\bibliography{main}

\end{document}